\documentclass[lettersize,journal,pdftex]{IEEEtran}
\IEEEoverridecommandlockouts                            
                       
\usepackage{graphics} % for pdf, bitmapped graphics files
\usepackage{epsfig} % for postscript graphics files
\usepackage{amsmath} % assumes amsmath package installed
\usepackage{amssymb}  % assumes amsmath package installed
\usepackage{graphicx}
\usepackage{cite}
\usepackage{picinpar}
\usepackage{tabularx}
\usepackage{url}
\usepackage{flushend}
\usepackage[latin1]{inputenc}
\usepackage{bigdelim}
\usepackage{colortbl}
\usepackage{soul}
\usepackage{multirow}
\usepackage{pifont}
\usepackage{color}
\usepackage{alltt}
\usepackage{enumerate}
\usepackage{siunitx}
\usepackage{breakurl}
\usepackage{epstopdf}
\usepackage{pbox}
\usepackage{comment}
\usepackage{bm}
\usepackage{booktabs}  
\usepackage{subfig}
\usepackage{stfloats} %????
\usepackage{algorithmic}
\usepackage{algorithm}
\usepackage{array}
\usepackage{textcomp}
\usepackage{verbatim}%????
\usepackage{float}
\usepackage{cancel}
\usepackage{diagbox}   
\usepackage{placeins}  % 导言区
\usepackage{afterpage}
\usepackage{stfloats}  % 放在导言区
\usepackage{hyperref}

\usepackage{xurl,cite,bm}
\usepackage{makecell}
\usepackage[table]{xcolor}
\usepackage{comment}

\newtheorem{remark}{Remark}

\definecolor{RowBlue}{RGB}{242,247,252}
 \definecolor{TextBlue}{RGB}{40,24,255}

\title{
DR-IPC:  {D}isturbance-{R}esilient  {I}ntegrated {P}lanning and {C}ontrol  for LiDAR-Based Quadrotor Navigation}

\author{Peng Liu$^{1}$, Jingyan Wang$^{1}$, Qipeng Ye$^{2}$, 
Wen Li$^{1}$, Jinya Su$^{1}$, \textit{Senior Member, IEEE},\\ Zuo Wang$^{1}$,  \textit{Member, IEEE}, Shihua Li$^{1}$, \textit{Fellow, IEEE}, Yunda Yan$^{3}$, \textit{Senior Member, IEEE}

\thanks{Corresponding author: Jinya Su} 

\thanks{$^{1}$School of Automation,  Key Laboratory of Measurement and Control of CSE, Ministry of Education, Institute of Intelligent Unmanned Systems, Southeast University, Nanjing 210096, China.
(e-mail: drlp@seu.edu.cn, jingyan.wang@seu.edu.cn,  lwen@seu.edu.cn,  sucas@seu.edu.cn, z.wang@seu.edu.cn, lsh@seu.edu.cn)
}

\thanks{$^{2}$School of Electrical Engineering and Automation, Xiamen University of Technology, Xiamen 361024, China.
(e-mail: iceteen@stu.xmut.edu.cn)}

\thanks{$^{3}$  Department of Computer Science, University College London, London, WC1E 6BT, UK.   
(e-mail: yunda.yan@ucl.ac.uk)}%
}

\begin{document}

\maketitle
\thispagestyle{empty}
\pagestyle{empty}

%%%%%%%%%%%%%%%%%%%%%%%%%%%%%%%%%%%%%%%%%%%%%%%%%%%%%%%%%%%%%%%%%%%%%

\begin{abstract}
LiDAR-based quadrotor navigation in cluttered environments remains challenging under external disturbances, particularly when obstacle-aware motion generation and disturbance-rejection control are handled in separate layers. 
This article presents disturbance-resilient integrated planning and control (DR-IPC), which combines lightweight path guidance with nonlinear model predictive control (NMPC) to directly generate angular velocity and thrust. 
An interconnected extended Kalman filter and nonlinear disturbance observer jointly provide filtered state estimates and reconstructed disturbances for NMPC prediction. 
The resulting formulation unifies nonlinear quadrotor dynamics, actuator constraints, local motion generation, and penalized safe-flight-corridor residuals without requiring a separate trajectory-optimization stage.
%
% Under bounded state- and disturbance-estimation errors, reference mismatch, and SFC residuals, a finite-horizon robustness analysis establishes prediction, optimal-value, and near-optimality bounds.
%
Gazebo and MARSIM simulations, together with indoor and outdoor experiments, validate DR-IPC under wind, suspended payloads, narrow passages,  ball impacts  and reactive avoidance of a dynamic obstacle.
In multi-goal navigation with disturbances, DR-IPC increases the number of completed missions from 1/10 to 9/10
% increases the success rate from 1/10 to 9/10 
in Gazebo and reduces the altitude RMSE from 0.34 to 0.01~m in experiments.
The complete system operates onboard at 100 Hz.
% and its implementation will be released as open source. Videos and supplementary materials are available at: \url{https://drpp316.github.io/DR-IPC-Page/}.
Supplementary videos are available on the project page  \url{https://drpp316.github.io/DR-IPC-Page/}, and the source code will be released.

%
%Hardware experiments further demonstrate reliable flight under payload variations, wind, impacts, and moving obstacles. 
%

\end{abstract}
% and the source code 


 \begin{figure}[ht]
    \centering
\includegraphics[scale=1.1]{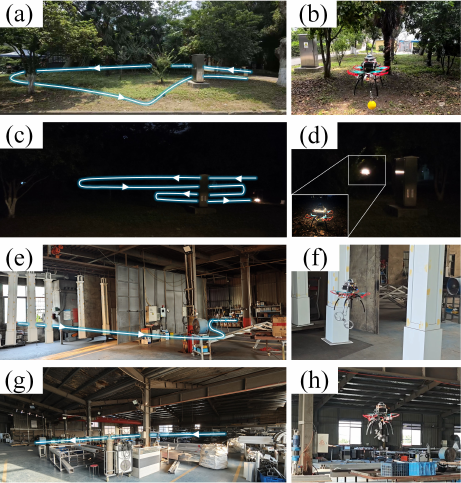}
 % \caption{{Outdoor multi-target navigation with disturbances}. {\footnotesize (a) takeoff, (b)-(c) in-flight snapshots, (d) experimental scene and  targets.}}
 \caption{{Real-world validation} of the proposed disturbance-resilient navigation in cluttered environments. {\footnotesize (a)--(d) flight routes and snapshots in the outdoor wooded scene during daytime and nighttime; (e)--(h) flight routes and snapshots in two real-life factory scenarios.}}
\label{outdoor1}
\end{figure}

% \begin{table}[h]
% \section*{Nomenclature}
%     % 1. 强制使用正常字号
%     \normalsize  
    
%     % 2. 调整行高 (字号变大后，1.3 就够了，1.4可能会太稀疏)
%     \renewcommand{\arraystretch}{1.3} 
%     % 3. 稍微减小列间距
%     \setlength{\tabcolsep}{3pt}       
%     % 4. 调整列宽：
%     % 0.12 + 0.15 + 0.70 = 0.97 < 1.0 (预留一点空间给 tabcolsep)
%     % 这样保证双栏排版下不会撑破边界
%     \begin{tabular}{@{} p{0.08\columnwidth} p{0.08\columnwidth} p{0.84\columnwidth} @{}}
    
%     {${ \mathbf{x}}_{k}$,} & {$\mathbf{y}_{k}$} & {Actual state and measured state vectors.} \\ 
    
%     {$\hat{ \mathbf{x}}^-_{k}$,} & {$\hat{ \mathbf{x}}^+_{k}$} & {A priori and a posteriori state estimates.} \\ 
    
%     {$\mathbf{d}_k$,} & {$\hat{\mathbf{d}}_k$} & {Actual lumped disturbance and its estimation.} \\ 

%     %{$\mathbf{\tilde{x}}_{k}$,} & {$\tilde{\mathbf{d}}_{k}$} & {State and disturbance estimation errors.} \\ 
    
%     {$\mathbf{P}_k^-$,} & {$\mathbf{P}_k^+$} & {A priori/posteriori error covariance matrices.} \\ 
    
%     {$\mathbf{Q}$,} & {$\mathbf{R}$} & {Process/measurement noise covariance matrices.} \\ 
    
%     {$\mathbf{K}_k$,} & {$\mathbf{\Lambda}$} & {Kalman gain and observer gain matrices.}\\ 
    
%      $\mathbf{x}_{i|k},$ & $\mathbf{x}^{r}_{i|k}$ & {$i$-th predictive state and its reference at time $k$.}\\

%     $\mathbf{u}_{i|k},$ & $\mathbf{u}^{r}_{i|k}$ & {$i$-th decision variable and its reference at time $k$.}\\

%     %{$\bm{\Phi}$,} & {$\bm{\Gamma}$} & {Discretized system state transition/input matrices.} \\
%     \end{tabular}
% \end{table}

\section{INTRODUCTION}
\IEEEPARstart{P}{roviding} onboard 3-D mapping and odometry,
%\IEEEPARstart{L}{}iDAR-based 
LiDAR-based quadrotors have become an important platform for autonomous navigation in cluttered and partially unknown environments \cite{ren2025survey},  % kwon2019robust
including inspection \cite{liu2025lidar,marshall2025redundant}, search and rescue\cite{lim2024autonomous}, %petrlik2020robust
precision agriculture \cite{su2023ai}, %lin2021quality
and aerial transportation \cite{sun2025agile}.
However, LiDAR maps contain only geometric information and do not capture wind, payload-induced forces, or modeling errors. Odometry noise further degrades the state measurements used for feedback \cite{tian2015mechatronic}. Consequently, these disturbances and  uncertainties can drive the LiDAR-based vehicle away from the nominally collision-free path. 

Most LiDAR-based quadrotor systems use a hierarchical planning and control (HPC) architecture \cite{ren2025survey,marshall2021survey}, where a planner generates collision-free motion from the LiDAR map and a downstream controller handles tracking 
%and disturbance rejection 
separately \cite{zhou2020ego}. 
This separation makes it difficult to adjust collision-avoidance motion and control actions together when disturbances cause path deviations. 
Integrated planning and control (IPC) reduces this gap by jointly optimizing local motion and control commands within a predictive backend  \cite{liu2025slope, liu2023integrated, wu2025flying}. 
However, existing LiDAR-based IPC methods provide only partial integration and limited (i.e., passive instead of active) robustness to external disturbances and noisy state measurements.

\begin{figure*}[tpb!]
    \centering
    \includegraphics[scale=0.39]{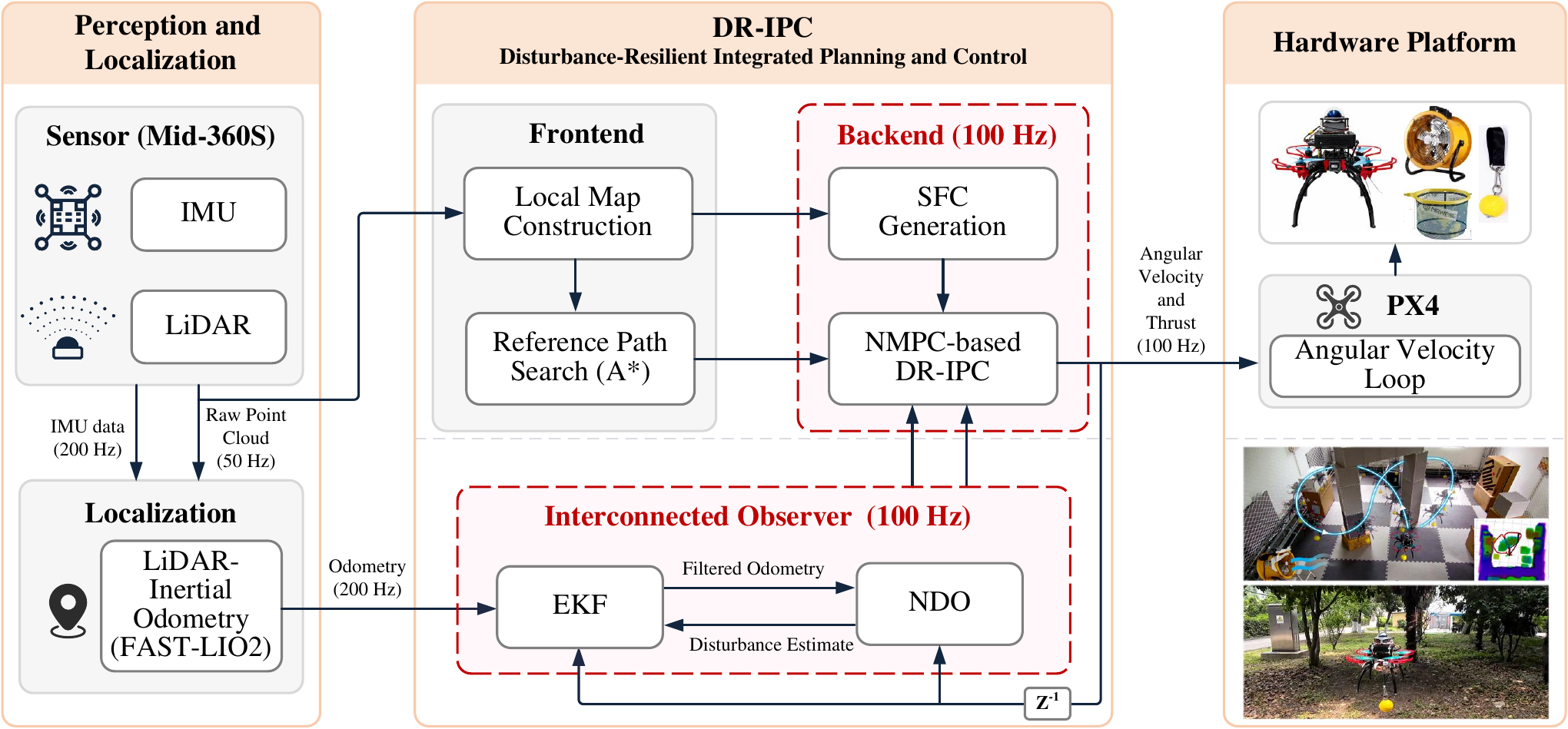}
    \caption{{Pipeline of the proposed DR-IPC}. {\footnotesize The  frontend provides  collision-free geometric guidance, the EKF+NDO supplies filtered states and reconstructed disturbances, and the NMPC backend jointly optimizes local motion and angular-velocity/thrust commands with SFC penalties and input constraints. The main contributions are highlighted in red.}}
    \label{pipeline}
\end{figure*}

To this end, on top of the classical IPC \cite{liu2023integrated}, this paper presents \textbf{DR-IPC}, a \textbf{D}isturbance-\textbf{R}esilient  \textbf{I}ntegrated \textbf{P}lanning and \textbf{C}ontrol algorithm for LiDAR-based quadrotor navigation. Unlike the original IPC, DR-IPC uses the reconstructed disturbance not only for feedback compensation but also within the NMPC prediction to jointly modify collision-avoidance motion and control inputs. The pipeline is shown in Fig.~\ref{pipeline}. 
%和下面创新点总结部分 重复
% A* provides lightweight global geometric guidance from the LiDAR map, while the safe-flight corridor (SFC) represents local free space as a convex region \cite{liu2023integrated}. A disturbance-aware nonlinear model predictive control (NMPC) backend then generates local motion, angular velocity, and thrust while accounting for nonlinear quadrotor dynamics, actuator constraints,  disturbances, and SFC violation penalties. An interconnected extended Kalman filter (EKF) and nonlinear disturbance observer (NDO) provide filtered state estimates and reconstruct lumped disturbances from noisy measurements. 
The main contributions are:

\begin{itemize}
  \item[(1)] \textbf{DR-IPC Algorithm}: A DR-IPC algorithm is developed for LiDAR-based quadrotor navigation in cluttered, noisy, and disturbance-rich environments. The A* path references, soft SFC penalties, nonlinear quadrotor dynamics, input constraints, and reconstructed disturbances are jointly handled in a unified NMPC backend.

  \item[(2)]  \textbf{Nonlinear Interconnected  Observer}:   A nonlinear interconnected EKF+NDO is designed for simultaneous state filtering and disturbance reconstruction in the presence of noise, providing noise-attenuated state estimates and smoother disturbance estimates for NMPC.

  % \item[(3)] \textbf{Soft-SFC Feasibility Enhancement}: We introduce SFC violation  penalties in the cost function to improve online feasibility while penalizing predicted corridor violations.

  \item[(3)] \textbf{Comparative Multi-Scenario Validation}:   DR-IPC  is validated in Gazebo and MARSIM, together with indoor and outdoor experiments covering wind disturbances, suspended-payload transportation, 
  narrow-passage navigation, impacts from a throwing ball, reactive avoidance of a dynamic obstacle, wooded and factory flights.
\end{itemize}

%$\mathbb{R}^{n \times m}$ denotes the set of real matrices with dimension $n \times m$. $\mathbb{R}^n$ and ${\mathbb{R}_{ \ge 0}}$ denote  the   $n$-dimensional Euclidean space  and  the set of non-negative real numbers, respectively.  $\mathbb{N}_+$ represents  the set of positive integers.  $\bm{I}$  is the identity matrix, and $\rho (\bm{M})$ represents the spectral radius of matrix $ \bm{M}$.  The  operator  $\mathrm{sgn}(\cdot)$ is the sign function. For discrete-time signals, the subscript $k$ denotes the value at $t=kT$, i.e., $\circ_k \triangleq \circ(k T)$, where $T$ is the sampling period. $\bm{\mathcal{O}} (T) $ is referred to be of the order of $T$.

\textbf{Notations:}
$\mathbb R^n$ and $\mathbb R^{n\times m}$ denote the $n$-dimensional
Euclidean space and the set of $n\times m$ real matrices, respectively.
$\mathbf{I}$ and $\mathbf{0}$ denote identity and zero matrices of appropriate dimensions, respectively. The unit-quaternion set is
$\mathbb S^3:=\{\mathbf q\in\mathbb R^4\mid\|\mathbf q\|=1\}$.
$\mathcal N(\boldsymbol\mu,\boldsymbol\Sigma)$ denotes the Gaussian
distribution with mean $\boldsymbol\mu$ and covariance
$\boldsymbol\Sigma$.
%$\Delta t$ is the sampling period.
For a symmetric matrix $\mathbf M\succeq\mathbf0$,
$\|\mathbf z\|_{\mathbf M}^{2}:=\mathbf z^\top\mathbf M\mathbf z$, while
$\|\mathbf z\|$ denotes the Euclidean norm. For a nonnegative integer $n$,
$\mathbb I_n:=\{0,\ldots,n\}$. Vector inequalities are interpreted
element-wise.

%tmp: Some new ref  \cite{wang2019tie}. exp ref \cite{Wang2024RAL}.

\begin{figure}[b]
    \centering
    \includegraphics[scale=0.99]{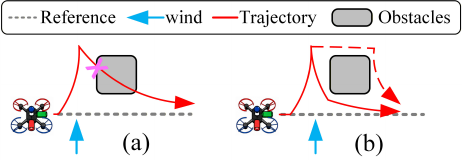}
    \caption{{Planning and control} under disturbances.
    %{\footnotesize (a) Hierarchical scheme with collision risk; (b) Integrated scheme with obstacle avoidance.}}
    {\footnotesize
   (a) in the HPC, the planner uses obstacle information, whereas the controller does not; disturbance-induced deviations therefore may  result in collision;
    (b) IPC uses obstacle information to jointly adjust local motion and control actions (the dashed and solid curves represent two feasible collision-avoidance routes).}}
    \label{hpc_ipc}
\end{figure}

\section{RELATED WORK}

\subsection{Obstacle-Aware Motion Generation and Predictive Control}

\subsubsection{Hierarchical Planning and Control}
Conventional quadrotor navigation separates path planning, trajectory optimization, and tracking control. A front-end planner typically uses a geometric or simplified point-mass model, while a back-end optimizer converts the path into a smooth nominal trajectory. Representative methods include SFC-based polynomial optimization \cite{liu2017planning}, minimum-snap generation  \cite{mellinger2011minimum}, EGO-Planner \cite{zhou2020ego}, and hierarchical two-layer MPC \cite{benders2025embedded}. A lower-level feedback controller then tracks the optimized trajectory.
%
%This modular architecture is efficient, but planning generally ignores external disturbances, whereas the tracker has no obstacle information and cannot revise the trajectory. Planning errors or insufficient disturbance rejection may therefore erode safety margins and cause collisions.

In this architecture, obstacle avoidance is handled only by the planner, typically using a simplified model without disturbance effects. 
The resulting path may be geometrically collision-free but not dynamically feasible under disturbances. 
At the control level, as illustrated in Fig.~\ref{hpc_ipc}(a), the tracking controller does not have obstacle information and continues to reduce the tracking error after the vehicle deviates from the reference. 
It may drive vehicles toward nearby obstacles without timely replanning. 
Consequently, the  performance of HPC may degrade in cluttered environments under disturbances.

 \begin{figure*}[t!]
    \centering
    \includegraphics[scale=1.48]{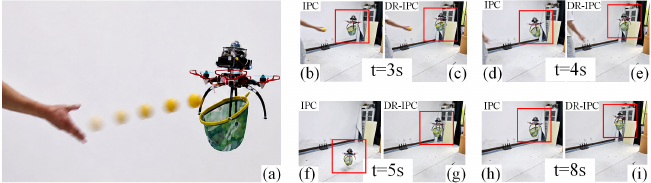}
     \caption{{Ball-impact hover experiment}. {\footnotesize (a) experimental setup;
    (b)--(i) IPC and DR-IPC transient responses at $t=3$, 4, 5, 8 s, respectively.}}
    \label{figure_ball2}
\end{figure*}

\begin{table*}[t]
    \centering
    \caption{Capability comparison of  representative quadrotor 
    %planning and control 
    methods. 
    ({\footnotesize$\checkmark$: yes; $\times$: no; --: not applicable or not reported.})}
    \label{tab:related_work_comparison}
    \scriptsize
    \setlength{\tabcolsep}{1.8pt}
    \renewcommand{\arraystretch}{1.10}
    \begin{tabular*}{\textwidth}{@{\extracolsep{\fill}}lccccccc@{}}
        \toprule
        \textbf{Method} & \textbf{Function} & \textbf{Architecture} &
        \textbf{\makecell{Obstacle-aware\\ planning}} &
        \textbf{\makecell{Estimated disturbance \\in prediction}} &
        \textbf{\makecell{Nonlinear\\MPC model}} &
        \textbf{\makecell{Cluttered\\real-world flight}} &
        \textbf{\makecell{Onboard\\perception}} \\
        
        \midrule
        Minimum-snap  \cite{mellinger2011minimum}
        & Planning and control  & Hierarchical & $\times$ & $\times$ & -- & $\checkmark$ & $\times$ \\
        
        EGO-Planner \cite{zhou2020ego}
        & Planning  & Hierarchical & $\checkmark$ & $\times$ & -- & $\checkmark$ & Camera \\

       Obstacle-aware NMPC \cite{lindqvist2020nmpc}
        & Planning and control & Integrated & $\checkmark$ & $\times$ & $\checkmark$ & $\times$ & $\times$ \\

        CBF-MPC \cite{ali2024mpccbf}
        & Planning and control & Integrated & $\checkmark$ & $\times$ & $\times$ & $\times$ & $\times$ \\
        
        MPCC \cite{romero2022model}
        & Planning and control & Integrated & $\times$ & $\times$ & $\checkmark$ & $\times$ & $\times$ \\

        IPC \cite{liu2023integrated}
        & Planning and control & Integrated & $\checkmark$ & $\times$ & $\times$ & $\checkmark$ & LiDAR \\

        Robust visual-servo NMPC \cite{zhang2021robust}
        & Tracking & -- & $\times$ & $\times$ & $\checkmark$ & -- & Camera \\

        Efficient tracking NMPC \cite{wang2021efficient}
        & Tracking & -- & $\times$ & $\times$ & $\checkmark$ & -- & $\times$ \\

        Embedded HMPC \cite{benders2025embedded}
        & Planning and control & Hierarchical & $\checkmark$ & $\times$ & $\checkmark$ & $\checkmark$ & $\times$ \\

        DA-MPC \cite{kim2024disturbance}
        & Tracking & -- & $\times$ & $\checkmark$ & $\checkmark$ & $\times$ & $\times$ \\

        DR-MPC \cite{li2025drmpc}
        & Tracking & -- & $\times$ & $\checkmark$ & $\checkmark$ & -- & Camera \\

        DOB-based tracking/landing \cite{Shao2021tmech,zhang2026tie}
        & Tracking & -- & $\times$ & $\checkmark$ & -- & -- & $\times$ \\

         \color{TextBlue}{\textbf{DR-IPC (ours)}}
        &  \color{TextBlue}{ \textbf{Planning and control}} 
        &  \color{TextBlue}{ \textbf{Integrated} }
        &  \color{TextBlue}{$\checkmark$} 
        &  \color{TextBlue}{$\checkmark$}
        &  \color{TextBlue}{$\checkmark$} 
        &  \color{TextBlue}{$\checkmark$}
        &  \color{TextBlue}{\textbf{LiDAR}} \\
        \bottomrule
    \end{tabular*}
\end{table*}

\subsubsection{Integrated Planning and Control}
Integrated schemes have attracted increasing attention  \cite{ren2025survey,marshall2021survey}. 
%  It therefore acts as a local planner-controller rather than a separate trajectory optimizer and tracker.  
It acts as a local predictive planner-controller that uses global guidance and obstacle information \cite{liu2023integrated}. 
%At each receding-horizon update, it optimizes collision-aware local motion and the control variables, rather than tracking a fixed local trajectory.
At each receding-horizon update, it jointly optimizes collision-aware local motion and control inputs, rather than merely tracking a prescribed trajectory without accounting for obstacles \cite{reiter2026temporal}.
As shown in Fig.~\ref{hpc_ipc}(b), when wind drives the vehicle away from the reference, the integrated schemes can adjust both the local motion around the obstacle and the corresponding control action.

Existing integrated methods couple motion generation and control in different ways.
Obstacle-aware NMPC \cite{lindqvist2020nmpc} imposes a prescribed  UAV--obstacle distance as a hard constraint. 
% Control barrier function (CBF)-
CBF-based MPC \cite{ali2024mpccbf} embeds forward-invariant safety conditions into MPC for collision avoidance.
% Model predictive contouring control (MPCC)  
MPCC \cite{romero2022model} jointly optimizes path progress and control for racing but omits obstacle avoidance. 
IPC \cite{liu2023integrated} combines A*  with SFC-constrained linear MPC, with subsequent extensions to  challenging terrain and dynamic environments \cite{liu2025slope,wu2025flying}.
However, its linear triple-integrator model derives  angular velocity and thrust through differential flatness, while integral thrust adaptation provides only limited disturbance compensation.
IPC \cite{liu2023integrated} therefore remains partially integrated and offers limited active disturbance-rejection capability.
Moreover, its hard SFC constraint may make online optimization infeasible or induce aggressive corrective motion after disturbance-induced deviations \cite{wang2021efficient}.
Overall, these representative methods rely mainly on nominal prediction and do not incorporate reconstructed disturbances into obstacle-aware motion optimization.

\subsection{Robust and Disturbance-aware MPC}

\subsubsection{Robust MPC} 
Robust MPC handles bounded  disturbances and uncertainties without explicitly reconstructing them.
Robust visual-servo NMPC \cite{zhang2021robust} maintains target visibility through constraint tightening. Efficient NMPC  \cite{wang2021efficient} converts inequality constraints into cost penalties and imposes a contraction constraint for real-time stable tracking. 
%In both cases, MPC remains a tracker of a prescribed trajectory or visual target. The robustness design neither reconstructs the disturbance nor couples disturbance rejection with obstacle-aware motion generation.
%These methods improve constrained tracking but do not use disturbance estimates for obstacle-aware motion generation.

\subsubsection{Disturbance-aware MPC} Disturbance observer  (DOB)-based quadrotor controllers have been developed for disturbance-rejection tracking and landing \cite{Shao2021tmech,zhang2026tie}, following the general disturbance-estimation and compensation framework \cite{chen2004disturbance,chen2016doboverview}.
Observer-based MPC further incorporates disturbance estimates into prediction. DA-MPC \cite{kim2024disturbance} injects an NDO estimate into the prediction model.  DR-MPC \cite{li2025drmpc} combines a generalized extended-state observer with predictive visual servoing for UAV pipeline inspection. 

Both branches employ MPC  for prescribed trajectory or visual-target tracking. Robust MPC does not reconstruct the disturbance, whereas disturbance-aware MPC incorporates its estimate into prediction. However, neither couples disturbance rejection with obstacle-aware local motion generation.

\subsection{Positioning of DR-IPC}
 
% The literature above reveals a gap between obstacle-aware motion generation and disturbance-aware control. Representative obstacle-aware methods omit online disturbance reconstruction, whereas disturbance-aware MPC methods exclude obstacle information from prediction. Consequently, neither branch uses reconstructed disturbances to jointly adapt collision-avoiding local motion and control actions. Table~\ref{tab:related_work_comparison} summarizes this distinction.
As summarized in Table~\ref{tab:related_work_comparison}, existing obstacle-aware methods generally omit online disturbance reconstruction, whereas disturbance-aware MPC methods, which mainly serve as low-level control strategies, do not include obstacle information in prediction. {The proposed DR-IPC bridges this gap by integrating disturbance reconstruction with obstacle-aware predictive control.} 
Building on our previous linear KF+DOB architecture \cite{liu2026interconnected}, the interconnected  observer is extended to nonlinear quaternion dynamics. 
The resulting EKF+NDO supplies filtered states and reconstructed disturbances to an obstacle-aware NMPC that jointly optimizes local motion, angular velocity and thrust.
Without a separate trajectory-optimization stage, the onboard system supports LiDAR-based navigation and suspended-payload transportation in disturbance-rich, cluttered outdoor and factory environments.

% The resulting EKF+NDO supplies filtered states and reconstructed disturbances to an obstacle-aware NMPC that jointly optimizes local motion, angular velocity and thrust.

\section{Modeling and Problem Formulation}

\subsection{Dynamic Modeling}

\begin{figure}[t]
    \centering
    \includegraphics[scale=0.38]{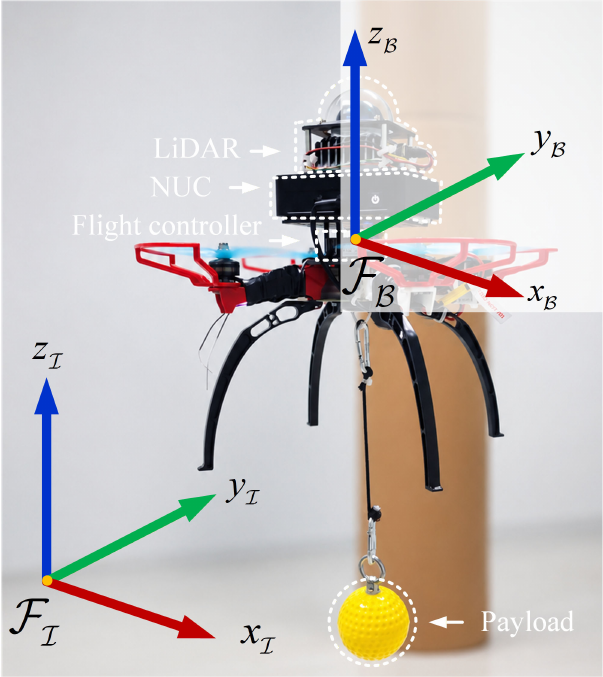}
    \caption{{Quadrotor platform and coordinate frames}.}
    \label{figurelabel2}
\end{figure}

%As shown in Fig. \ref{figurelabel2}, two coordinate frames are used: the inertial frame $\mathcal{F_I}=\{O_{\mathcal{I}},x_{\mathcal{I}},y_{\mathcal{I}},z_{\mathcal{I}}\}$ and the body-fixed frame $\mathcal{F_B}=\{O_{\mathcal{B}},x_{\mathcal{B}},y_{\mathcal{B}},z_{\mathcal{B}}\}$ attached to the quadrotor center of mass.
As shown in Fig.~\ref{figurelabel2}, the quadrotor is described in the
inertial frame $\mathcal {F_I}$ and the body-fixed frame $\mathcal {F_B}$, which
is attached to its center of mass.
The position $\mathbf{p}\in\mathbb{R}^{3}$, velocity $\mathbf{v}\in\mathbb{R}^{3}$, and gravity $\mathbf{g}=[0,\,0,\,g]^\top\in\mathbb{R}^{3}$  are expressed in $\mathcal{F_I}$.
The angular velocity  $\mathbf{w}\in\mathbb{R}^{3}$ is expressed in $\mathcal{F_B}$.
The unit quaternion $\mathbf{q}=[q_w,q_x,q_y,q_z]^\top\in\mathbb{S}^{3}\subset\mathbb{R}^{4}$ represents the attitude, and the total thrust $T\in\mathbb{R}$ acts along the body $z_{\mathcal{B}}$-axis.
 
%The vector $\boldsymbol{\Phi}(\mathbf{q})$ maps the body $z_{\mathcal{B}}$-axis into $\mathcal{F_I}$, such that $\boldsymbol{\Phi}(\mathbf{q})T$ is the thrust force expressed in the inertial frame. 

\begin{figure*}[t]
    \centering
    \includegraphics[scale=0.9]{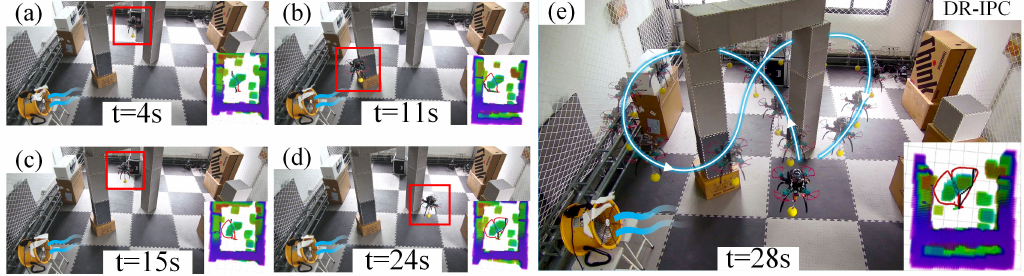}
    \caption{{Indoor multi-waypoint navigation} with disturbances.  {\footnotesize  
    (a)--(e) DR-IPC transient responses at $t=4$, 11, 15, 24, 28~s, respectively.}}
    \label{figure_indoor1}
\end{figure*}

Define the  state   as $\mathbf{x} = [\mathbf{p}^\top,\, \mathbf{q}^\top,\, \mathbf{v}^\top]^\top \in \mathbb{R}^{10}$, the control input   as $\mathbf{u} = [\mathbf{w}^\top,\,T]^\top \in \mathbb{R}^4$ and the UAV mass  as $m$. The  quadrotor dynamics are  described by
\begin{equation}
    \dot{\mathbf{x}} = f(\mathbf{x}, \mathbf{u}) + \mathbf{B} \mathbf{d},
    \label{eq:compact_model}
\end{equation}
where  $\mathbf{d}\in\mathbb{R}^{3}$ is the lumped acceleration disturbance in $\mathcal F_I$, accounting for external forces and unmodeled dynamics. The nonlinear dynamics $f$ and  disturbance input matrix $\mathbf B$ are
\begin{equation}
    f(\mathbf{x}, \mathbf{u}) =
       \begin{bmatrix}
        \mathbf{v} \\
        \dfrac{1}{2}\boldsymbol{\Omega}(\mathbf{q}) \mathbf{w} \\
        -\mathbf{g} + \dfrac{\boldsymbol{\Phi}(\mathbf{q})}{m} T \\
    \end{bmatrix}, ~
    \mathbf{B} =
      \begin{bmatrix}
        \mathbf{0}_{3 \times 3} \\
        \mathbf{0}_{4 \times 3} \\
        \mathbf{I}_{3 \times 3}
    \end{bmatrix}.
\end{equation}
The   thrust direction map and quaternion kinematics matrix are  
\begin{equation*}
    \boldsymbol{\Phi}(\mathbf{q}) =
    \begin{bmatrix}
        2(q_x q_z + q_w q_y) \\
        2(q_y q_z - q_w q_x) \\
        1 - 2(q_x^2 + q_y^2)
    \end{bmatrix}, ~
    \boldsymbol{\Omega}(\mathbf{q}) =
    \begin{bmatrix}
        -q_x & -q_y & -q_z \\
        \ \    q_w  & -q_z &  \ \   q_y \\
        \ \    q_z  &  \ \   q_w & -q_x \\
        -q_y &  \ \    q_x &  \ \   q_w
    \end{bmatrix}.
\end{equation*}

Considering process noise, the discrete-time dynamics are  
\begin{equation}
\begin{aligned}
 { \mathbf{x}}_{k+1} & =  
 \mathbf{F}(\mathbf{x}_k,\mathbf{u}_k,\mathbf{d}_k)+\boldsymbol{\eta}_k,\\
 \mathbf{F}(\mathbf{x}_k,\mathbf{u}_k,\mathbf{d}_k)&= { \mathbf{x}}_{k} +\Delta t \, [f(\mathbf{x}_k, \mathbf{u}_k) + \mathbf{B} \mathbf{d}_k],
 \end{aligned}
\end{equation}
where $\Delta t$ is the sampling period and noise $\boldsymbol{\eta}_k\sim\mathcal{N}(\mathbf{0},\mathbf{Q})$. The LiDAR-inertial odometry provides the measurement  
\begin{equation}
    \mathbf{y}_{k}
    %=[\mathbf{p}_{m,k}^{\top},\mathbf{q}_{m,k}^{\top},\mathbf{v}_{m,k}^{\top}]^{\top}
    =
\mathbf{H}\mathbf{x}_{k}+\boldsymbol{\nu}_{k},
    \label{eq:ekf_measurement_model}
\end{equation}
where $\mathbf{H}=\mathbf{I}_{10} $ and  $\boldsymbol{\nu}_{k}\sim \mathcal{N}(\mathbf{0}, \mathbf{R} )$ is  the measurement noise.

\begin{comment}
\begin{table*}[htbp!]
\centering
\caption{Performance Comparison of Control Architectures}
\label{tab:comparison}
\definecolor{RowBlue}{RGB}{242,247,252}
\definecolor{TextBlue}{RGB}{40,24,255}
\setlength{\tabcolsep}{4pt}        % ↓ 减小列间距
\renewcommand{\arraystretch}{1.2} % ↓ 减小行高
\small
\newcolumntype{Y}{>{\centering\arraybackslash}X}
\begin{tabularx}{0.95\textwidth}{lYYYYY}
\toprule[1.1pt]
\textbf{Methods} & \textbf{Rate} & \textbf{\makecell{Matched\\Dist.}} & \textbf{\makecell{Unmatched\\Dist.}} & \textbf{Disturbance} & \textbf{Constraints} \\ 
\midrule
NMPC + $\mathcal{L}_1$ & Single & $\checkmark$ (slow) & $\times$ & Const / Slow & High Risk \\ 
Offset-free NMPC (Case 1) & Single & $\checkmark$ (slow) & $\checkmark$ & Const / Slow & Guaranteed \\ 
\rowcolor{RowBlue}
\color{TextBlue}{MR-NMPC+DOB (Case 2)} & Multi & $\checkmark$ (fast) & $\times$ & Wide-band & High Risk \\ 
\rowcolor{RowBlue}
\color{TextBlue}{MR-NMPC+DOB (Case 3)} & Multi & $\checkmark$ (fast) & $\checkmark$ & Wide-band & Low Risk \\ 
\bottomrule[1.1pt]
\end{tabularx}
\end{table*}
\end{comment}

\subsection{A* Guidance and Safe-Flight Corridor}
Using the local occupancy map, A* generates a collision-free geometric path and replans it when a newly detected obstacle blocks the  path. The   reference path at update $k$ is  
\begin{equation}
    \mathcal{R}_k = \{\mathbf{p}^r_{0|k}, \mathbf{p}^r_{1|k}, \dots, \mathbf{p}^r_{N|k}\},
\label{eq:reference_path}
\end{equation}
where  $\mathbf{p}^r_{i|k}\in\mathbb{R}^3$ is the position reference at step $i$.  
Since $\mathcal{R}_k$ is generated geometrically and guides the local motion, it is not required to satisfy the quadrotor dynamics. %Based on the local map,
The SFC is constructed from an occupancy map after obstacle inflation and defines a local collision-free convex region  
\begin{equation}
    \mathcal{C}_{k}=\{\mathbf{p}\in\mathbb{R}^3 \mid \mathbf{A}_{k}\mathbf{p}\leq \mathbf{b}_{k}\},
    \label{eq:sfc}
\end{equation}
%where $\mathbf{A}_{k}\in\mathbb{R}^{n_{\mathrm{s}}\times3}$, $\mathbf{b}_{k}\in\mathbb{R}^{n_{\mathrm{s}}}$  and $n_{\mathrm{s}}$ is the SFC num.
where $\mathbf A_k\in\mathbb R^{n_{\mathrm{s}}\times3}$,
$\mathbf b_k\in\mathbb R^{n_{\mathrm{s}}}$, and
$n_{\mathrm{s}}$ is constraint number.

\subsection{Problem Statement}

At update $k$, DR-IPC receives the noisy odometry measurement $\mathbf y_k$, the A*-based geometric guidance $\mathcal R_k$, and the local SFC $\mathcal C_k$. Since the true state $\mathbf x_k$ and lumped disturbance $\mathbf d_k$ are unavailable, online state filtering and disturbance
reconstruction are required for predictive control.

%The SFC is used to guide obstacle avoidance but is not imposed as a hard constraint. 

% In DR-IPC,
% The SFC is incorporated as an obstacle-avoidance penalty rather than a hard constraint \cite{wang2021efficient}. 
% Disturbances and abrupt map updates may temporarily drive the  motion outside the SFC, making a hard-constrained optimal control problem (OCP) infeasible.
% Since obstacles are inflated before SFC construction, a brief violation does not necessarily indicate a physical collision.
Sudden disturbances or map updates may drive the  motion outside the SFC. A hard SFC constraint may make the optimal control problem (OCP) infeasible or induce aggressive recovery and oscillatory flight.
% A soft SFC is thus preferable \cite{wang2021efficient}, but its violation must be quantified and embedded in the OCP.
The challenge is therefore to use a soft penalty for SFC violations while maintaining OCP feasibility under disturbances \cite{wang2021efficient}.

Accordingly, the objective is to jointly optimize dynamically feasible local motion and the angular velocity/thrust  using the filtered state and reconstructed disturbance, while considering nonlinear dynamics, soft SFC penalties, and input constraints.

%\section{DR-IPC FRAMEWORK}
\section{DR-IPC Algorithm}
\subsection{Nonlinear Interconnected Observer}
%for State Filtering and Disturbance Estimation } \label{DisturbanceEstimation}

To estimate the state and lumped disturbance from noisy odometry, the linear KF+DOB architecture in \cite{liu2026interconnected} is extended to the nonlinear quaternion model. The EKF uses the reconstructed disturbances in its state prediction, while the reduced-order NDO uses the posterior attitude and velocity from the EKF.
This modular interconnection enables joint state filtering and disturbance reconstruction and allows either estimator to be upgraded or disabled.
%This interconnection provides filtered state and reconstructed disturbance for the NMPC backend while retaining a modular estimator structure.
%Let $\hat{ \mathbf{x}}_{k}$ and $\mathbf{P}_{k}$ denote the EKF state estimate and covariance. 
%The EKF update is  
The EKF recursion is given by
%Let $\hat{ \mathbf{x}}_{k}$ and $\mathbf{P}_{k}$ denote the EKF state estimate and covariance. 
\begin{comment}   nominal EKF
 \begin{equation}
    \left\{\begin{matrix}
     \begin{aligned}
       \hat{ \mathbf{x}}^-_{k+1}  &=  \hat{ \mathbf{x}}^+_{k} +\Delta t\, f_o( \hat{ \mathbf{x}}^+_{k},\mathbf{u}_{k},\hat{\mathbf{d}}_{k}),\\
    \bar{\mathbf{P}}_{k+1}    &= \mathbf{A}_{k}\mathbf{P}_{k}\mathbf{A}_{k}^{\top}+\mathbf{Q}_{k},\\
    \mathbf{K}_{k+1}    &= \bar{\mathbf{P}}_{k+1}\mathbf{H}^{\top}
\left(\mathbf{H}\bar{\mathbf{P}}_{k+1}\mathbf{H}^{\top}
    +\mathbf{R}_{k}\right)^{-1},\\
    \hat{ \mathbf{x}}^+_{k+1}    &= \hat{ \mathbf{x}}^-_{k+1}
    +\mathbf{K}_{k+1}    \left(\mathbf{y}_{k+1}-\mathbf{H}\hat{ \mathbf{x}}^+_{k+1}\right),\\
    \mathbf{P}_{k+1}    &= \left(\mathbf{I}-\mathbf{K}_{k+1}\mathbf{H}\right)\bar{\mathbf{P}}_{k+1},
     \end{aligned}
\end{matrix}\right.
\label{eq:ekf_update}
\end{equation}
\end{comment}

 \begin{equation}
    \left\{\begin{matrix}
     \begin{aligned}
       \hat{ \mathbf{x}}^-_{k}  &=  \mathbf{F}( \hat{ \mathbf{x}}^+_{k-1},\mathbf{u}_{k-1},\hat{\mathbf{d}}_{k-1}),\\
    {\mathbf{P}}^-_{k}    &= \mathbf{J}_{k-1}\mathbf{P}^+_{k-1}\mathbf{J}_{k-1}^{\top}+\mathbf{Q},\\
    \mathbf{K}_{k}    &=  {\mathbf{P}}^-_{k}\mathbf{H}^{\top} \left(\mathbf{H}\mathbf{P}^-_{k}\mathbf{H}^{\top}
    +\mathbf{R}\right)^{-1},\\
    \tilde{\mathbf{x}}^+_{k}    &= \hat{ \mathbf{x}}^-_{k}     +\mathbf{K}_{k}    \left(\mathbf{y}_{k}-\mathbf{H}\hat{ \mathbf{x}}^-_{k}\right),\\
     \hat{\mathbf{x}}^+_{k}     &=\left[(\tilde{\mathbf p}_{k}^{+})^\top,     \frac{(\tilde{\mathbf q}_{k}^{+})^\top}
    {\|\tilde{\mathbf q}_{k}^{+}\|},     (\tilde{\mathbf v}_{k}^{+})^\top     \right]^\top,\\
    \mathbf{P}^+_{k}    &= \left(\mathbf{I}-\mathbf{K}_{k}\mathbf{H}\right)\mathbf{P}^-_{k},
     \end{aligned}
\end{matrix}\right.
\label{eq:ekf}
\end{equation}
where $\hat{\mathbf x}_k^-$ and $\hat{\mathbf x}_k^+$ are the prior and posterior state  estimates, respectively. 
$\mathbf P_k^-$ and $\mathbf P_k^+$ are their error covariance  matrices. $\mathbf K_k$ is the Kalman gain.
The intermediate estimate $\tilde{\mathbf x}_{k}^+$ is obtained before quaternion normalization, which ensures $\hat{\mathbf q}_{k}^+\in\mathbb S^3$ \cite{baritzhack1991quaternion,choukroun2006novel}. 
$\mathbf{J}_k=\partial\mathbf F/\partial\mathbf x|_{(\hat{\mathbf x}_k^+,\mathbf u_k,\hat{\mathbf d}_k)}$ is the Jacobian matrix, and $\hat{\mathbf d}_k$ is supplied by the NDO in \eqref{eq:NDOB}.

Following  \cite{chen2004disturbance}, %chen2016novel
the   lumped acceleration disturbance is assumed to be constant or slowly varying. For generic disturbances, please refer to \cite{su2016relationship} and the references therein. The discrete-time reduced-order NDO is 
 \begin{comment}
  %exogenous signal with $\dot{\boldsymbol{\xi}}=\mathbf{0}$ and $\mathbf{d}=\boldsymbol{\xi}$.
%By choosing the nonlinear auxiliary function $\mathbf{p}_{o}(\mathbf{x})=\mathbf{\Lambda}\hat{\mathbf{v}}^+$,
\begin{equation}
\label{eq:ndo_update}
    \left\{\begin{matrix}
     \begin{aligned}
    \dot{\mathbf{z}} &= -\mathbf{\Lambda}\left(-\mathbf{g}+\frac{\boldsymbol{\Phi}(\hat{\mathbf{q}}^+)}{m}T+\mathbf{z}+\mathbf{\Lambda}\hat{\mathbf{v}}^+ \right),\\
    \hat{\mathbf{d}} &= \mathbf{z}+\mathbf{\Lambda}\hat{\mathbf{v}}^+,
     \end{aligned}
\end{matrix}\right.
\end{equation}
where $\mathbf{z}\in\mathbb{R}^{3}$ is the internal NDO state and $\mathbf{\Lambda}$ is the observer gain.
In discrete implementation, \eqref{eq:ndo_update} is integrated by forward Euler:
 \end{comment}
\begin{equation}
\left\{\begin{matrix}
\begin{aligned}
    \mathbf{z}_{k}    &=    \mathbf{z}_{k-1}    -    \Delta t\,\mathbf{\Lambda}    \left(-\mathbf{g}+\frac{\boldsymbol{\Phi}(\hat{\mathbf{q}}^+_{k-1})}{m}T_{k-1}+\hat{\mathbf{d}}_{k-1}\right),\\
    \hat{\mathbf{d}}_{k}    &=    \mathbf{z}_{k}+\mathbf{\Lambda}\hat{\mathbf{v}}^+_{k},
\end{aligned}
\end{matrix}\right.
    \label{eq:NDOB}
\end{equation}
where $\mathbf{z}_k\in\mathbb{R}^{3}$ is the internal observer state and $\mathbf{\Lambda}\in\mathbb{R}^{3\times3}$ is the observer gain. The $\hat{\mathbf{q}}^+_k$ and $\hat{\mathbf{v}}^+_{k}$ are extracted
from $\hat{\mathbf x}^+_k$ in \eqref{eq:ekf}.  As shown in Fig.~\ref{pipeline}, the disturbance estimate
corrects the EKF prediction, while the filtered state drives the NDO update. 
The pair $(\hat{\mathbf x}^+_{k},\hat{\mathbf d}_{k})$ is then supplied to the NMPC.

%The interconnection structure reduces the sensitivity of disturbance estimation to noisy raw odometry and prevents the NMPC backend from relying on a purely nominal prediction model.
%The interconnection structure reduces the sensitivity of disturbance estimation to noisy raw odometry and prevents the NMPC backend from relying on a purely nominal prediction model.

\subsection{Disturbance- and Obstacle-Aware NMPC}

At update $k$, $\mathbf x_{i|k}$ and $\mathbf u_{i|k}$ denote the state and input predicted at step $i$. 
% Let $\mathbb X\subseteq\mathbb R^3\times\mathbb S^3\times\mathbb R^3$ denote the admissible state set.
%For an $N$-step horizon, t
The NMPC optimizes the input sequence
\begin{equation}
\begin{aligned}
    % \mathbf U_k&=\{\mathbf u_{i|k}\}_{i=0}^{N-1}, \ 
        \mathbf U_k&=\{\mathbf u_{0|k},\cdots , \mathbf u_{N-1|k}\},\\
    \mathbf u_{i|k}&=
    [\mathbf w_{i|k}^\top,T_{i|k}]^\top,\ 
    i\in\mathbb I_{N-1}.
\end{aligned}
\label{eq:u_sequence}
\end{equation}
 The admissible input set is
\begin{equation}
    \mathbb U=
    \Big\{ [\mathbf w^\top,T]^\top\  \Big| \ 
   |\mathbf w|\preceq\mathbf w_{\max},
    T_{\min}\leq T\leq T_{\max}\Big\}.
\end{equation}

\begin{comment}
 The input constraint set is
\begin{equation}
    \mathbb{U}=\{\mathbf{u}\mid \mathbf{u}_{\min}\leq\mathbf{u}\leq\mathbf{u}_{\max}\}.
\end{equation}
\end{comment}

The   reference   is   $\mathbf{x}^{r}_{i|k}= [(\mathbf{p}_{i|k}^{r})^\top ,(\mathbf{q}_{i|k}^{r})^\top,(\mathbf{v}_{i|k}^{r})^{\top}]^\top$, with $\mathbf u^r_{i|k}=[\mathbf 0_3^\top,mg]^\top$.
$\mathbf{p}_{i|k}^{r}$ is provided by A* in \eqref{eq:reference_path}, $\mathbf{v}_{i|k}^{r}$ is computed by differences of $\mathbf{p}_{i|k}^{r}$, and $\mathbf q^r_{i|k}$ is level with yaw aligned with the path.  Under \eqref{eq:sfc}, the SFC violation residual is  
\begin{equation}
\boldsymbol{\epsilon}_{i|k}
    =
[\mathbf{A}_{k}\mathbf{p}_{i|k}-\mathbf{b}_{k}]_{+},
  %  \quad \boldsymbol{\epsilon}_{i|k}\in\mathbb{R}^{n_{\mathrm{s}}}_{\geq0},
   \quad   i\in\mathbb I_{N-1},
\label{eq:sfc_soft_residual}
\end{equation}
where $[h]_+ =\max\{0,h\}$ is the element-wise positive part and $[h]_+ =0$ implies  the SFC constraint is satisfied \cite{lindqvist2020nmpc,wang2021efficient,zhou2022swarm}.   The stage cost, terminal cost, and finite-horizon cost are
\begin{gather*}
    %\label{stage_cost_analysis}
    \ell_{i|k}=
    \|\mathbf{x}_{i|k}-\mathbf{x}^{r}_{i|k}\|_{\mathbf{Q}_w}^2
    +\|\mathbf{u}_{i|k}-\mathbf{u}^{r}_{i|k}\|_{\mathbf{R}_w}^2
+\|\boldsymbol{\epsilon}_{i|k}\|_{\mathbf{S}_w}^2,\\
    V_{f}=\|\mathbf{x}_{N|k}-\mathbf{x}^{r}_{N|k}\|_{\mathbf{P}_w}^2,\quad
    \mathcal J_k=\sum_{i=0}^{N-1}\ell_{i|k}+V_{f},
    %\label{terminal_cost_analysis}
\end{gather*}
where $\mathbf Q_w,\mathbf P_w\in\mathbb R^{10\times10}$,
$\mathbf R_w\in\mathbb R^{4\times4}$, and
$\mathbf S_w\in\mathbb R^{n_{\mathrm{s}}\times n_{\mathrm{s}}}$ are
symmetric positive-definite weighting matrices.

\begin{comment}
For small attitude deviations and continuously sign-consistent unit
quaternions, the Euclidean quaternion difference is locally equivalent to the
  attitude error $\delta\boldsymbol{\phi}_{i|k}$:
\begin{equation}
    \|\mathbf q_{i|k}-\mathbf q^r_{i|k}\|^2
    \approx
    \frac{1}{4}
    \|\delta\boldsymbol{\phi}_{i|k}\|^2 .
\end{equation}
The quaternion term in
$\|\mathbf x_{i|k}-\mathbf x^r_{i|k}\|_{\mathbf Q_w}^2$ provides a locally
positive-definite measure of the attitude-tracking error.
\end{comment}

At each update, the NMPC backend solves \cite{Hanover2021RAL,Jackson2021RAL}
\begin{subequations}
\label{eq:dripc_nmpc}
\begin{align}
    % V_k&=\min_{\mathbf{U}_k}\mathcal J_k\\ 
     %
     \mathbf U_k^\star & {=\arg\min_{\mathbf{U}_k}\mathcal J_k}\\
     \text{s.t.}\quad  \mathbf{x}_{0|k}&=\hat{\mathbf{x}}^+_k, \label{ocp_ini}\\ 
    \mathbf{d}_{i|k}&=\hat{\mathbf{d}}_k,  &&i\in\mathbb I_{N-1},\\
     { \mathbf{x}}_{i+1|k}  &=   
    \mathbf{F}(\mathbf{x}_{i|k}, \mathbf{u}_{i|k},\mathbf{d}_{i|k}),  &&i\in\mathbb I_{N-1},\label{eq:nmpc_dynamics_constraint}\\
     \boldsymbol{\epsilon}_{i|k}
    &= [ \mathbf{A}_{k}\mathbf{p}_{i|k}-\mathbf{b}_{k}]_{+},  &&i\in\mathbb I_{N-1},\\
    %
      % \mathbf{x}_{i|k}  &\in \mathbb{X},    &&i\in\mathbb I_{N},\\
     %
      \mathbf{u}_{i|k} &\in \mathbb{U},    &&i\in\mathbb I_{N-1}.
\end{align}
\end{subequations}
After solving \eqref{eq:dripc_nmpc}, the first optimal input is directly sent to the flight controller:     $\mathbf{u  }_k=\mathbf{u}_{0|k}^\star.$
For clarity, the online DR-IPC procedure is summarized in Algorithm~\ref{alg:dripc}.

\begin{algorithm}[t]
\caption{Online execution of DR-IPC}
\label{alg:dripc}
\begin{algorithmic}[1]
\REQUIRE LiDAR cloud $\mathcal P_k$, odometry $\mathbf y_k$,  goal $\mathbf p_g$, and RC command $E_k$
\ENSURE Angular-velocity and thrust command $\mathbf u_k$
\STATE Initialize the EKF, NDO, local map, A*, and NMPC.
\WHILE{autonomous flight is enabled by $E_k$}
    \STATE Update the local occupancy map $\mathcal M_k$ using $\mathcal P_k$.
    \STATE Update the EKF using $\mathbf y_k$, $\mathbf u_{k-1}$, and $\hat{\mathbf d}_{k-1}$ to obtain $\hat{\mathbf x}_k^+$.
    \STATE Update the NDO using $\hat{\mathbf x}_k^+$ and $\mathbf u_{k-1}$ to obtain $\hat{\mathbf d}_k$.
    \IF{a new goal is received or the remaining A* path is obstructed}
        \STATE Replan the A* path from $\hat{\mathbf p}_k^+$ to $\mathbf p_g$ on $\mathcal M_k$.
        % \STATE Reset the NMPC warm start.
    \ENDIF
    \STATE Generate the horizon reference $\mathcal R_k$ from the  A* path.
    \STATE Construct the SFC $ \mathcal{C}_{k}$ around $\mathcal R_k$ using $\mathcal M_k$.
    \STATE Solve \eqref{eq:dripc_nmpc} using $\hat{\mathbf x}_k^+$, $\hat{\mathbf d}_k$, $\mathcal R_k$, and $ \mathcal{C}_{k}$.
    % \IF{the solver succeeds}
    %     \STATE Apply the first optimal input $\mathbf u_k=\mathbf u_{0|k}^{\star}$.
    % \ELSE
    %     \STATE Apply the safe-hover command $\mathbf u_k=\mathbf u_{\mathrm{safe}}$.
    % \ENDIF
    \STATE Apply the first optimal input $\mathbf u_k=\mathbf u_{0|k}^{\star}$.
\ENDWHILE
\end{algorithmic}
\end{algorithm}

\begin{remark} 
The soft SFC penalty maintains OCP feasibility when disturbances or map updates temporarily drive the predicted motion outside the SFC. Moreover, with obstacle inflation, a small SFC violation does not necessarily imply physical collision, while the soft penalty avoids the aggressive corrective motion.
% DR-IPC incorporates SFC violations as soft penalties, permitting brief deviations under disturbances or map updates. With obstacle inflation, a small SFC violation does not necessarily indicate physical collision.
% SFC violations are penalized in $\mathcal J_k$ rather than imposed as hard constraints. 
% With obstacle inflation, brief violations may remain within the safety margin without collision.
\end{remark}

\begin{remark}
The disturbance estimate is held constant over the prediction horizon. If a disturbance model is available, this zero-order prediction can be replaced by an exosystem for  disturbance preview and compensation \cite{su2016relationship,zhang2025PREDMPPI}.
%The NDO reconstructs the lumped  disturbance, which is held constant over  NMPC horizon. If the disturbance dynamics are known or identifiable, this zero-order model can be replaced by an exosystem, $\dot{\boldsymbol{\xi}}=\mathbf{A}_{d}{\boldsymbol{\xi}}$ and $\mathbf{d}=\mathbf{C}_{d}{\boldsymbol{\xi}}$.
%This extends DR-IPC from compensating unknown constant disturbances to predicting and compensating disturbances with modelable dynamics over the NMPC horizon.
\end{remark}   

% \begin{remark}
% % To avoid quaternion sign ambiguity in the cost, signs are chosen consistently such that $(\mathbf q^r_{i|k})^\top\mathbf q_{i|k}\geq0$.
% Let $\theta_{i|k}$ denote the relative rotation angle. For unit quaternions,
% \begin{equation}
% \|\mathbf q_{i|k}-\mathbf q^r_{i|k}\|^2
% = 4\sin^2\!\left(\frac{\theta_{i|k}}{4}\right)
% = \frac{\theta_{i|k}^2}{4} +\mathcal O(\theta_{i|k}^4).
% \end{equation}
% Hence, the quaternion term in the cost locally measures the attitude error \cite{Hanover2021RAL,Jackson2021RAL}.
% \end{remark}

\begin{figure}[b]
    \includegraphics[scale=0.6]{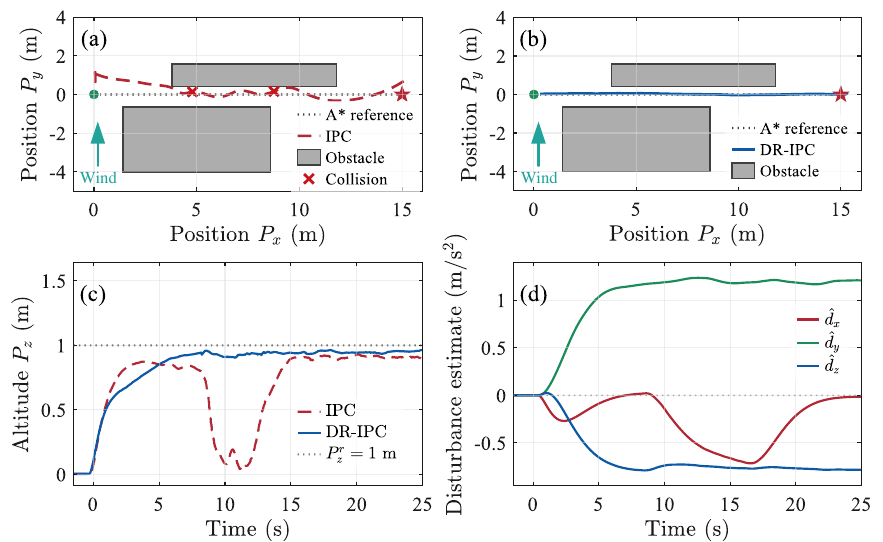}
     \caption{{Narrow-passage navigation} under constant wind. { \footnotesize (a) IPC trajectory;
    (b) DR-IPC trajectory; (c) altitude responses with 
    $p_z^r=1$~m; (d) DR-IPC disturbance estimates.}}
    \label{sim1-constant-wind}
    \end{figure}

\section{Comparative Simulation}

All simulations are conducted on a PC equipped with an Intel i7-14700F CPU, an NVIDIA GeForce RTX 4060Ti GPU, and 32~GB of RAM. 
%The software configuration is summarized in Table~\ref{tab:software_setup}.
The software configuration is summarized in
Table~\ref{tab:software_setup}, and the main  parameters are listed in   Table~\ref{tab:validation_parameters}.
DR-IPC is compared with the state-of-the-art IPC method
\cite{liu2023integrated}.
 
\begin{comment}
\begin{figure}[h]
    \centering
    \includegraphics[scale=0.3]{fig/gazebo/sim0/sim_environment0.pdf}
    \caption{Navigation in Narrow-Passage.}
    \label{figurelabel-gazebo1}
\end{figure}
\end{comment} 

 \begin{table}[t]
    \centering
    \caption{Software implementation of DR-IPC.}
    \label{tab:software_setup}
    \small
    \begin{tabularx}{\columnwidth}{lX}
        \toprule
        \textbf{Module} & \textbf{Implementation} \\
        \midrule
        Operating system & Ubuntu 20.04 and ROS1 Noetic \\
        Localization & FAST-LIO2 \cite{xu2022fastlio2} \\
        Point-cloud input & Registered LiDAR  cloud from FAST-LIO2 \\
        Local mapping & ROG-Map-style occupancy grid \cite{ren2023rogmap} \\
        Path search & A* graph search \cite{hart1968formal} \\
        SFC construction  & Polyhedral SFC \cite{liu2017planning} \\
        % Predictive backend  & Disturbance-aware NMPC with penalized SFC residuals (proposed) \\
        % State/disturbance estimation &  Interconnected EKF+NDO\\
        NMPC solver & ACADO \cite{houska2011auto,houska2011acado} with qpOASES \cite{ferreau2014qpoases} \\
        Simulation & Gazebo 11 and MARSIM \cite{kong2023marsim} \\
        Flight stack & PX4 v1.13.0 \\
        \bottomrule
    \end{tabularx}
\end{table}

\subsection{Narrow-Passage Navigation under Constant Wind}
This simulation evaluates disturbance rejection in confined flight. 
The quadrotor  traverses a narrow passage under constant wind along the positive $y$-axis. 
In Fig.~\ref{sim1-constant-wind}(d), $\hat d_y$ captures the steady crosswind,  while $\hat d_z$ mainly reflects vertical thrust-to-mass mismatch. 
The transient  $\hat d_x$ captures model residuals during acceleration and deceleration.
As shown in Figs.~\ref{sim1-constant-wind}(a) and (b), IPC deviates from the A* reference and collides twice, whereas DR-IPC remains close to the reference and completes the passage without collision. 
The altitude histories in
Fig.~\ref{sim1-constant-wind}(c) show that IPC drops to $0.042$~m, while DR-IPC remains above $0.850$~m and reduces the altitude RMSE from $0.484$ to $0.075$~m.

\begin{table}[t]
    \centering
    \caption{Interconnected-observer ablation under wind and a suspended payload.  {\footnotesize($\downarrow$: lower is better; pp: percentage points.)}}
    \label{tab:changing_wind_observer_ablation}
    \small
    \setlength{\tabcolsep}{3.0pt}
    \begin{tabular}{@{}lccc@{}}
        \toprule
        \textbf{Metric} & \textbf{NDO} & \textbf{EKF+NDO} &
        \textbf
        {Reduction} \\
        \midrule
        {Disturbance-vector RMSE (m/s$^2$) $\downarrow$} & {0.260} & {0.036} & {86.1\%} \\
        Disturbance-rate RMS (m/s$^3$) $\downarrow$ & 11.6 & 6.1 & 47.3\% \\
        Angular velocity RMS (rad/s) $\downarrow$ & 0.81 & 0.45 & 44.4\% \\
        $w_y$ saturation ratio (\%) $\downarrow$ & 36.0\% & 2.1\% & 33.9~pp \\
        Attitude RMS ($^\circ$) $\downarrow$ & 11.3 & 8.6 & 23.7\% \\
        \bottomrule
    \end{tabular}
\end{table}

     \begin{figure}[t]
    \includegraphics[scale=0.6]{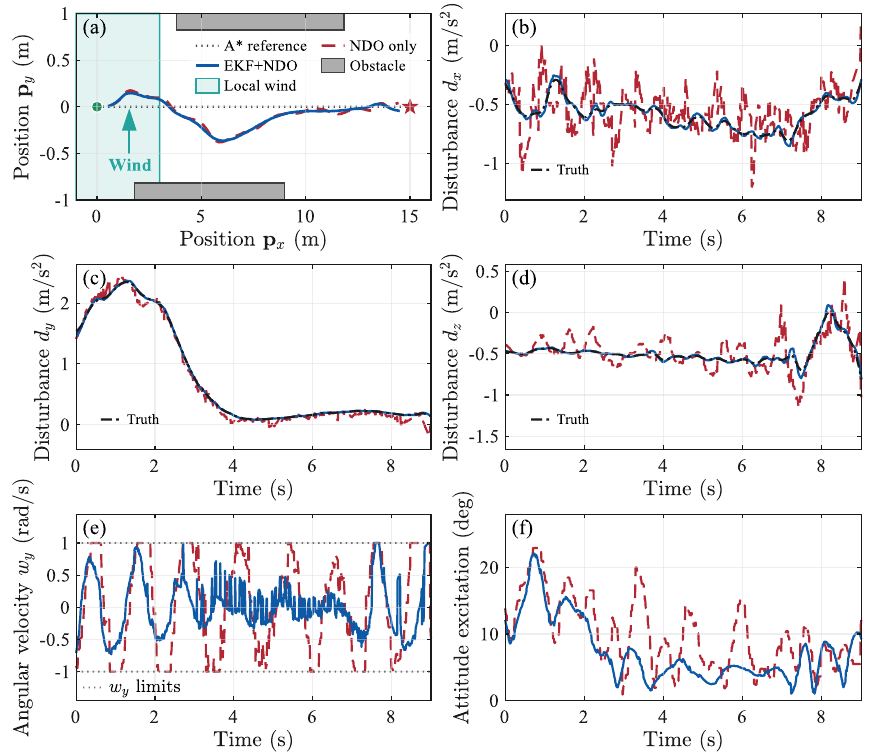}
\caption{{Interconnected observer ablation}.
     {\footnotesize
    (a) top-view trajectories and local wind region;
    (b)--(d) disturbance estimates $\hat d_x, \hat d_y, \hat d_z$ and ground truths;
    (e) angular velocity command $w_y$ and its limits;
    (f) attitude excitation.}}
    \label{sim1-changing-wind}
    \end{figure}

\subsection{Interconnected-Observer Ablation}
This simulation isolates the contribution of the interconnected EKF+NDO under a suspended payload and a local time-varying wind along the positive $y$-axis:
    $F_w(t)=2+\sin(0.8\pi t)\ \mathrm{N}$.
The two configurations use identical flight settings. In the NDO baseline, the NDO and NMPC directly use raw odometry. In  EKF+NDO, the EKF filters the states supplied to both NDO and NMPC, and the reconstructed disturbance is fed back into the EKF and NMPC prediction as described in Section IV-A.

% The quantitative results are summarized in  Table~\ref{tab:changing_wind_observer_ablation}. 
% The disturbance variation RMS characterizes the temporal fluctuation of the reconstructed disturbance. The angular velocity RMS reflects
% the overall fluctuation of the angular velocity commands, while the saturation ratio denotes the percentage of angular velocity approaching limits. The  attitude RMS quantifies the resulting attitude fluctuations in roll
% and pitch.

The quantitative results are summarized in Table~\ref{tab:changing_wind_observer_ablation}. 
The disturbance-vector RMSE measures 3-D disturbance estimation accuracy relative to the ground truth, while the disturbance-rate RMS quantifies fluctuations in the reconstructed disturbance.
The angular velocity RMS and $w_y$ saturation ratio characterize control-command activity and saturation, respectively. The attitude RMS quantifies the resulting roll--pitch fluctuations.
The responses are shown in Fig.~\ref{sim1-changing-wind}.
The disturbance ground truth is independently reconstructed from the Gazebo ground-truth states.

The similar trajectories in Fig. ~\ref{sim1-changing-wind}(a) indicate that the main improvements lie in disturbance estimation, control-command smoothness, and attitude stability rather than path geometry.
In Figs.~\ref{sim1-changing-wind}(b)--(d), EKF+NDO follows the ground-truth trends with fewer high-frequency fluctuations, reducing the {disturbance-vector RMSE from $0.260$ to $0.036$~m/s$^2$} and the {disturbance-rate RMS} by $47.3\%$. Consequently, the {angular velocity command RMS decreases from $0.81$ to $0.45$~rad/s}, and the $w_y$ saturation ratio drops from $36.0\%$ to $2.1\%$ in Fig.~\ref{sim1-changing-wind}(e). 
Fig.~\ref{sim1-changing-wind}(f) further shows that the {roll--pitch RMS} decreases from $11.3^\circ$ to $8.6^\circ$. 
These results demonstrate that the interconnected observer preserves accurate disturbance reconstruction while suppressing  noise, thereby reducing fluctuations in both the control commands and attitude response.

%  \begin{table*}[t]
%     \centering
%     \caption{DR-IPC parameters for simulation and experiments.}
%     \label{tab:validation_parameters}
%     \scriptsize
%     \setlength{\tabcolsep}{3pt}
%     \renewcommand{\arraystretch}{1.08}
%     \begin{tabular*}{\textwidth}{@{\extracolsep{\fill}}lccccccc@{}}
%         \toprule
%         \textbf{Case} &
%         $N$ & $\Delta t$ (ms) & $\mathbf w_{\max} $ (rad/s) & $T_{\max}$ (N) &
%         $\boldsymbol{\Lambda}$ & $\mathbf Q$ & $\mathbf R$ \\
%         \midrule
%         Simulation & $20$ & $50$ & $(1.0,1.0,1.5)$ &
%         $20.2$ & $2\mathbf I_3$ & $2\mathbf I_{10}$ & $\mathbf I_{10}$ \\
%         Experiment & $20$ & $50$ & $(1.0,1.0,1.5)$ &
%         $25.7$ & $\mathbf I_3$ & $\mathbf I_{10}$ & $5\mathbf I_{10}$ \\
%         \bottomrule
%     \end{tabular*}
%     \vspace{0.7mm}
%     \begin{tabular*}{\textwidth}{@{\extracolsep{\fill}}lcccc@{}}
%         \toprule
%         \textbf{Case} & $\mathbf Q_w$ & $\mathbf P_w$ &
%         $\mathbf R_w$ & $\mathbf S_w$ \\
%         \midrule
%         Simulation &
%         $\operatorname{diag}(45,45,100,8,5,5,8,4,4,8)$ &
%         $\operatorname{diag}(30,30,100,3,1,1,3,1,1,8)$ &
%         $\operatorname{diag}(12,12,12,1.5)$ & $\mathbf I_8$ \\
%         Experiment &
%         $\operatorname{diag}(45,45,80,8,5,5,8,4,4,8)$ &
%         $\operatorname{diag}(30,30,80,3,1,1,3,1,1,8)$ &
%         $\operatorname{diag}(12,12,12,2)$ & $\mathbf I_8$ \\
%         \bottomrule
%     \end{tabular*}
% \end{table*}

\begin{table*}[t]
    \centering
    \caption{DR-IPC parameters used in simulations and experiments.}
    \label{tab:validation_parameters}
    \scriptsize
    \renewcommand{\arraystretch}{1.08}

    \setlength{\tabcolsep}{5pt}
    \begin{tabular}{@{}lccccccc@{}}
        \toprule
        \textbf{Case} &
        $N$ &
        $\Delta t$ (ms) &
        $\mathbf w_{\max}$ (rad/s) &
        $T_{\max}$ (N) &
        $\boldsymbol{\Lambda}$ &
        $\mathbf Q$ &
        $\mathbf R$ \\
        \midrule
        Simulation &
        $20$ & $50$ & $(1.0,1.0,1.5)$ & $20.2$ &
        $2\mathbf I_3$ & $2\mathbf I_{10}$ & $\mathbf I_{10}$ \\
        Experiment &
        $20$ & $50$ & $(1.0,1.0,1.5)$ & $25.7$ &
        $\mathbf I_3$ & $\mathbf I_{10}$ & $5\mathbf I_{10}$ \\
        \bottomrule
    \end{tabular}

    \vspace{1.2mm}

    \setlength{\tabcolsep}{4pt}
    \begin{tabular}{@{}lcccc@{}}
        \toprule
        \textbf{Case} &
        $\mathbf Q_w$ &
        $\mathbf P_w$ &
        $\mathbf R_w$ &
        $\mathbf S_w$ \\
        \midrule
        Simulation &
        $\operatorname{diag}(45,45,100,8,5,5,8,4,4,8)$ &
        $\operatorname{diag}(30,30,100,3,1,1,3,1,1,8)$ &
        $\operatorname{diag}(12,12,12,1.5)$ &
        $\mathbf I_8$ \\
        Experiment &
        $\operatorname{diag}(45,45,80,8,5,5,8,4,4,8)$ &
        $\operatorname{diag}(30,30,80,3,1,1,3,1,1,8)$ &
        $\operatorname{diag}(12,12,12,2)$ &
        $\mathbf I_8$ \\
        \bottomrule
    \end{tabular}
\end{table*}

\begin{figure}[t]
    \centering
\includegraphics[scale=0.60]{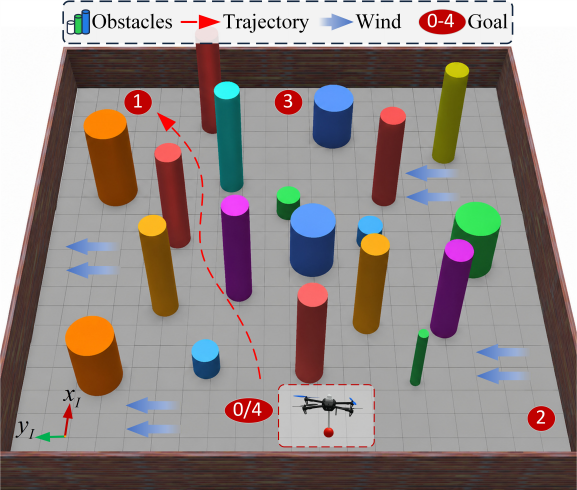}
    \caption{{Multi-goal Gazebo environment} with disturbances.}
    \label{figurelabel-gazebo2}
\end{figure}

\begin{table}[t]
    \centering
    \caption{
    %Multi-goal navigation performance over 10 trials.
    Multi-goal navigation over ten Gazebo trials. {\footnotesize(events: collisions or falls; average events: events in completed missions divided by the  completed missions; $\uparrow/\downarrow$: higher/lower is better; bold: best.)}}
    \label{tab:multigoal_navigation_results}
    \footnotesize
    \setlength{\tabcolsep}{2.0pt}
    \begin{tabular*}{\columnwidth}{@{\extracolsep{\fill}}lcccc@{}}
        \toprule
        \textbf{Method}& \textbf{{Disturbance}} &
         {\makecell{\textbf{Completed} \\\textbf{missions} $\uparrow$}} &
        {\makecell{\textbf{Reached}\\\textbf{goals} $\uparrow$}} &
        {\makecell{\textbf{Average}\\\textbf{events} $\downarrow$}} \\
        \midrule
        IPC     & $\times$  & \textbf{10/10} & \textbf{40/40} & 1   \\
        IPC     &  {$\checkmark$}  & 1/10  & 9/40  & 5   \\
        DR-IPC  &  {$\checkmark$}  & 9/10  & 37/40 & \textbf{0.1} \\
        \bottomrule
    \end{tabular*}
\end{table}

\subsection{Multi-Goal Navigation }

Three configurations are evaluated in Gazebo: IPC without disturbance, IPC with disturbance, and DR-IPC with disturbance. 
IPC without disturbances establishes nominal task feasibility, whereas IPC and DR-IPC are compared under identical disturbed conditions to evaluate disturbance resilience.
For the disturbed cases, the quadrotor is subjected to both constant wind and suspended-payload disturbances.
Each configuration is evaluated over 10 trials  in the environment shown in Fig.~\ref{figurelabel-gazebo2}. 
Each trial requires the vehicle to reach 4 goals sequentially, yielding $4\times10$ prescribed goals for each configuration. 
A goal is counted as reached when its horizontal distance is below $0.8$~m, and a mission is completed after all four goals are reached.
Minor collision and fall events are recorded and do not terminate a trial if the vehicle can continue the mission.  
%The reported event rate is averaged over completed missions.

As shown in Table~\ref{tab:multigoal_navigation_results}, IPC achieves full mission completion without disturbances but degrades markedly under wind and payload effects, completing only 1 of 10 missions and 9 of 40 goals. In contrast, DR-IPC completes 9 of 10 missions and reaches 37 of 40 goals under the same disturbances, while reducing average collision/fall events from 5 to 0.1 per completed mission. 
% These results demonstrate that disturbance-aware prediction significantly improves multi-goal navigation reliability under disturbances.
This comparison shows that DR-IPC
maintains higher mission completion with fewer
collision and fall events under combined wind and payload
disturbances.

\begin{figure}[t]
    \centering
\includegraphics[scale=0.35]{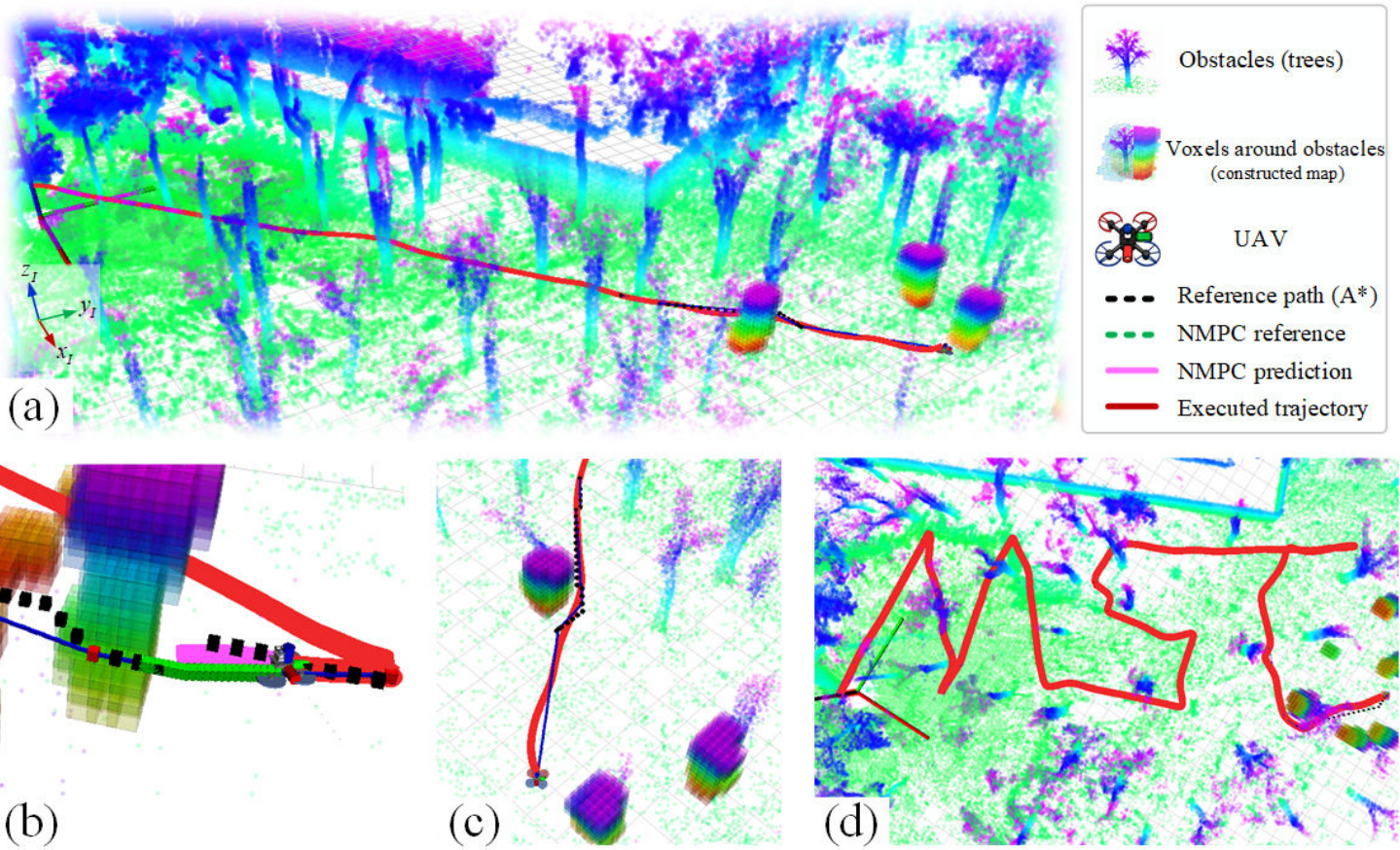}
 \caption{ {Navigation in the MARSIM forest environment}.
{\footnotesize 
{(a) overall navigation;}
{(b) local view I;}
{(c) local view II;}
{(d) multi-waypoint autonomous navigation.}
}}
\label{figurelabel-marsim}
\end{figure}

\subsection{Navigation in an Unstructured MARSIM Environment}
The complete DR-IPC pipeline is further evaluated in the unstructured 3-D forest environment of MARSIM to assess system-level integration and cross-simulator applicability. As shown in Fig.~\ref{figurelabel-marsim}, the A* reference, NMPC prediction, and executed trajectory are continuously updated during flight, demonstrating reliable online navigation in cluttered, unstructured environments beyond Gazebo.

% The complete DR-IPC pipeline is evaluated in the unstructured three-dimensional forest environment provided by MARSIM. Unlike the preceding controlled Gazebo studies, this simulation focuses on system-level integration and cross-simulator applicability.

% As shown in Fig.~\ref{figurelabel-marsim}, the A* reference, NMPC prediction,
% and executed trajectory are continuously updated during navigation through the
% cluttered forest. The results demonstrate that the proposed framework can be
% deployed beyond structured Gazebo environments for online navigation in
% unstructured three-dimensional scenes.

\begin{table}[t]
    \centering
    \caption{Hardware used in the experiments.}
    \label{tab:hardware_setup}
    \small
    \begin{tabularx}{\columnwidth}{@{}l>{\raggedright\arraybackslash}Xc@{}}
        \toprule
        \textbf{Component} & \textbf{Model}& \textbf{Mass (kg)} \\
        \midrule
        Onboard computer & ASUS NUC 13 (i5-1340P)  &\rdelim\}{8}{*}[\hspace{0.2em}1.705]\\
        %\rdelim\}{8}{*}[\makecell[l]{1.725 kg}]
        LiDAR & Livox Mid-360S \\
        Flight controller & CUAV V5 Nano& \\
        Battery & ACE Tattu  4S-2300 mAh& \\
        Frame & Totem Q250& \\
       Radio receiver & RadioLink R12DSE & \\
        Motor & Tmotor F60PRO KV2550& \\
        ESC & EMAX 45A& \\
         \midrule
        Suspended payload & Yellow ball & 0.225 \\
        Radio transmitter & RadioLink AT9S Pro & - \\
       Industrial fan & 350-2 (14 inch, 750 W) & - \\
        \bottomrule
    \end{tabularx}
\end{table}

\begin{figure}[t]
    \centering
    \includegraphics[scale=0.6]{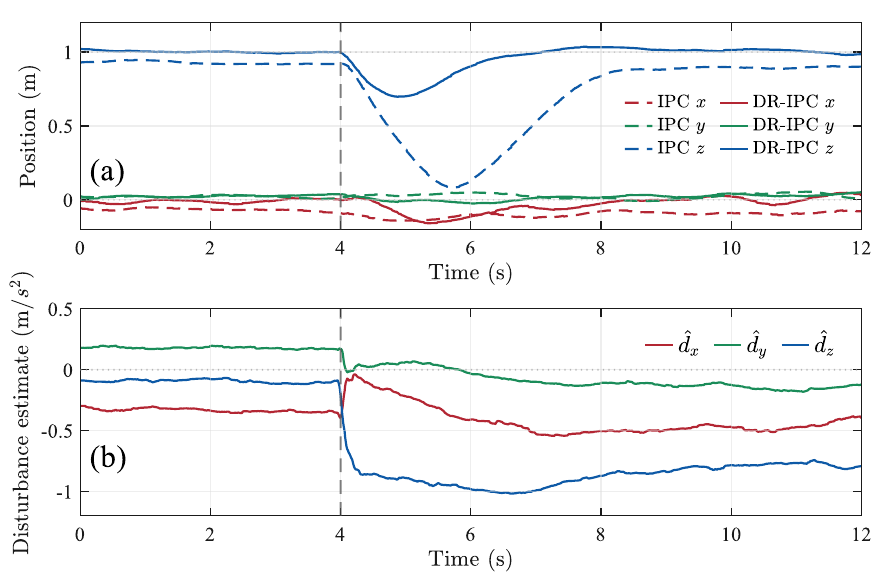}
    \caption{{Position responses and disturbance estimates}. {\footnotesize (a) position responses; (b) DR-IPC disturbance estimates.}}
    \label{ball_plot}
\end{figure}

\section{Experimental Validation}

The experimental platform and hardware components are summarized in Table~\ref{tab:hardware_setup}, while the corresponding control parameters are listed in Table~\ref{tab:validation_parameters}. In the comparative experiments, DR-IPC is evaluated against the state-of-the-art IPC method \cite{liu2023integrated} under comparable conditions.
 
\subsection{Ball-Impact Rejection in Hover}

This experiment evaluates the transient disturbance-rejection capability of DR-IPC by manually throwing the 0.225-kg yellow ball listed in  Table~\ref{tab:hardware_setup} into a suspended net beneath the hovering quadrotor. 
The net suspension length and ball-release position are fixed across all tests to provide approximately repeatable disturbances.
Experimental snapshots are shown in Fig.~\ref{figure_ball2}, with the reference hover position $\mathbf p^r=[0,0,1]^\top$~m. The corresponding position responses and disturbance estimates are shown in Fig.~\ref{ball_plot}.
%and the impact occurs at $t=4$~s.

As shown in Fig.~\ref{ball_plot}(a), the impact induces a significant altitude drop in IPC, with a minimum value of 0.08~m. 
DR-IPC effectively mitigates the disturbance, maintaining a minimum altitude of 0.70~m and returning to the $\pm0.05$~m altitude-error band within 2.30~s.  
Fig.~\ref{ball_plot}(b) shows that the reconstructed disturbance responds rapidly to the impact at $t=4$~s. The most pronounced variation occurs in $\hat d_z$, consistent with the altitude deviation in Fig.~\ref{ball_plot}(a).
These results demonstrate that DR-IPC rapidly reconstructs and compensates for the disturbance, substantially reducing altitude loss and restoring the commanded hover.

\subsection{Suspended-Payload Transportation with Wind}

%This subsection will test whether the disturbance-aware prediction model improves flight stability when unknown payload-induced forces are present. The comparison can include the original IPC, DR-IPC. Fig. \ref{figure_indoor1}. Fig. \ref{figure_indoor1_plot}.

\begin{figure}[t]
    \centering
    \includegraphics[scale=0.46]{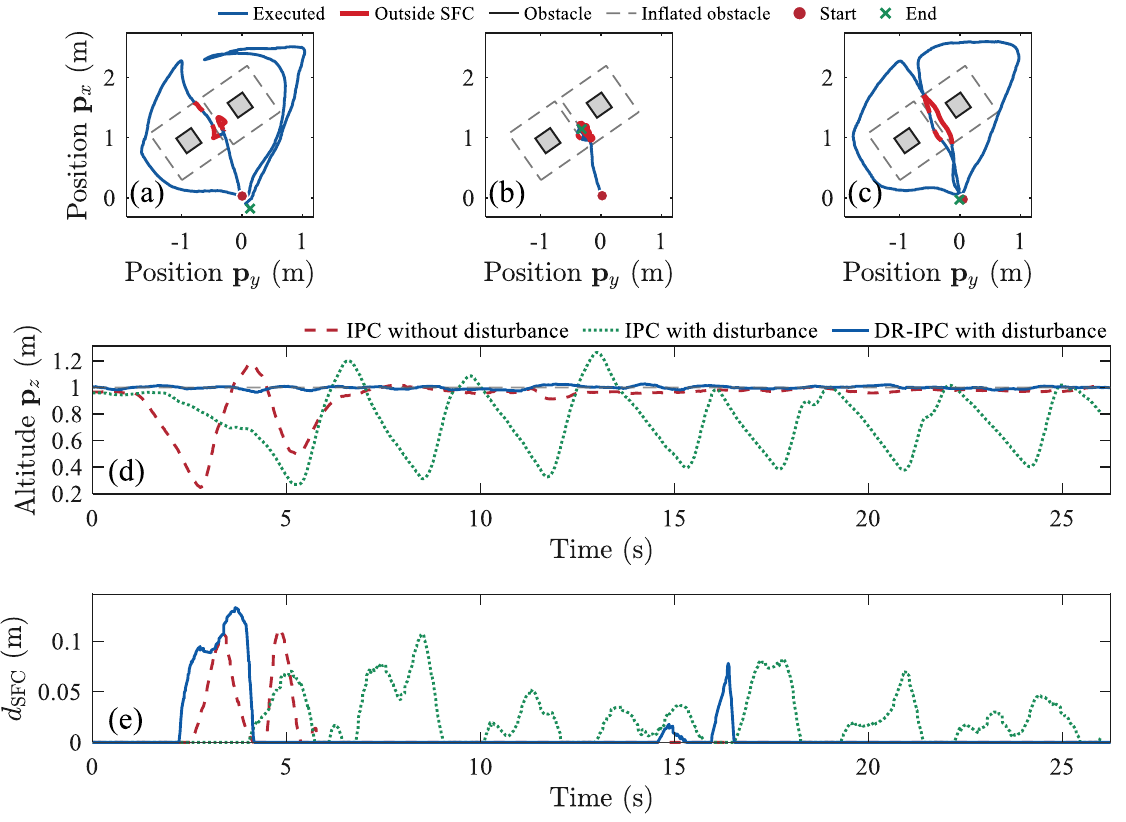}
\caption{{Navigation through an archway.}
{\footnotesize (a)--(c) top-view trajectories of IPC without disturbance, IPC with disturbance, and DR-IPC with disturbance, respectively, where red segments indicate penetration into the inflated obstacle regions; (d) altitude responses; (e) SFC-violation depth.}}
%The trajectory color denotes the altitude.
    \label{figure_indoor1_plot}
\end{figure}

This experiment evaluates whether disturbance-aware prediction and the soft SFC penalty can maintain stable narrow-archway navigation under combined wind and suspended-payload disturbances.
Experimental snapshots of DR-IPC are shown in Fig.~\ref{figure_indoor1}, with $p_z^r=1$~m.
Fig.~\ref{figure_indoor1_plot} compares the top-view trajectories, altitude responses, and SFC violations.
Fig.~\ref{figure_indoor1_plot}(e) reports the SFC-violation depth $d_{\mathrm{SFC}}$, where $d_{\mathrm{SFC}}>0$ indicates the trajectory is out of the SFC. 
The red segments in Figs.~\ref{figure_indoor1_plot}(a)--(c) indicate entry into the inflated obstacle region.
Table~\ref{tab:payload_navigation_performance} summarizes the route completion and altitude performance. 

Without disturbances, IPC eventually completes the passage, but the hard SFC constraint produces repeated corrective motion after the vehicle leaves the corridor, resulting in pronounced altitude oscillations, with $p_z^{\min}=0.25$~m and $\operatorname{RMSE}_{p_z}=0.17$~m.
Under wind and payload disturbances, the deviation becomes persistent. 
Although A* still provides a geometrically collision-free path, IPC cannot execute it while simultaneously satisfying the hard SFC, vehicle dynamics, and input constraints. 
Consequently, the out-of-SFC condition lasts $T_{\mathrm{out}}=$18.77~s and the vehicle fails to traverse the archway.

DR-IPC addresses both effects through disturbance-aware prediction and a soft SFC penalty. Disturbance compensation reduces the motion deviation, while the penalized SFC residual permits a brief  penetration into the inflated margin instead of treating it as a hard feasibility condition. DR-IPC therefore completes the passage with an out-of-SFC duration of $T_{\mathrm{out}}=$2.98~s, $p_z^{\min}=0.97$~m, and  $\operatorname{RMSE}_{p_z}=0.01$~m, without pronounced altitude oscillations.
This comparison shows that the soft penalty retains corridor guidance while avoiding the oscillatory corrective motion under disturbance.

\begin{table}[t]
    \centering
    \caption{ {Archway-navigation results.} {\footnotesize
   {(bold: best; $p_z^r=1$~m.)}}}
\label{tab:payload_navigation_performance}
    \footnotesize
    \setlength{\tabcolsep}{2.8pt}
    \renewcommand{\arraystretch}{1.05}
    \begin{tabular}{@{}lccccc@{}}
        \toprule
         {\textbf{Method}} &
         {\textbf{Disturbance}} &
       {\textbf{Passage}} &
        {\makecell{\boldmath$p_z^{\min}$\\\textbf{(m)}}} &
        {\makecell{\boldmath$\operatorname{RMSE}(p_z)$\\\textbf{(m)}}} &
        {\makecell{\textbf{Out-of-SFC time}\\\boldmath$T_{\mathrm{out}}$ \textbf{(s)}}} \\
        \midrule
        IPC    &$\times$  &$\checkmark$ & 0.25          & 0.17          & 3.13          \\
        IPC    &$\checkmark$ &$\times$  & 0.27          & 0.34          & 18.77         \\
        DR-IPC &$\checkmark$ &$\checkmark$ & \textbf{0.97} & \textbf{0.01} & \textbf{2.98} \\
        \bottomrule
    \end{tabular}
\end{table}

\subsection{Suspended-Payload Transportation with Dynamic Obstacle}

\begin{figure}[t]
    \centering
    \includegraphics[scale=0.98]{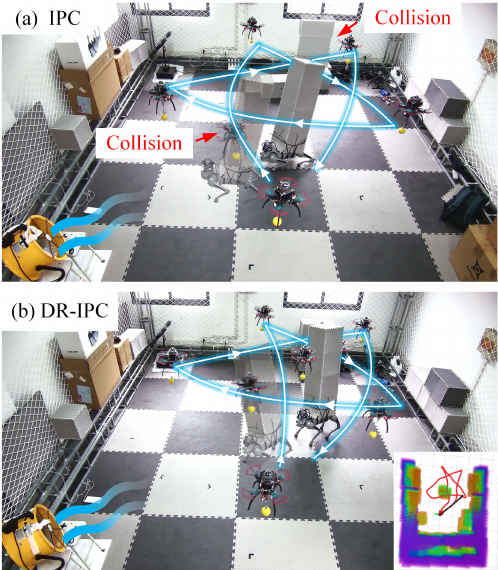}
    \caption{{Indoor multi-waypoint navigation} with disturbances and a dynamic obstacle.
    {\footnotesize (a) IPC; (b) DR-IPC.}}
    \label{figure_indoor3}
\end{figure}

\begin{figure}[t]
    \centering
    \includegraphics[scale=0.6]{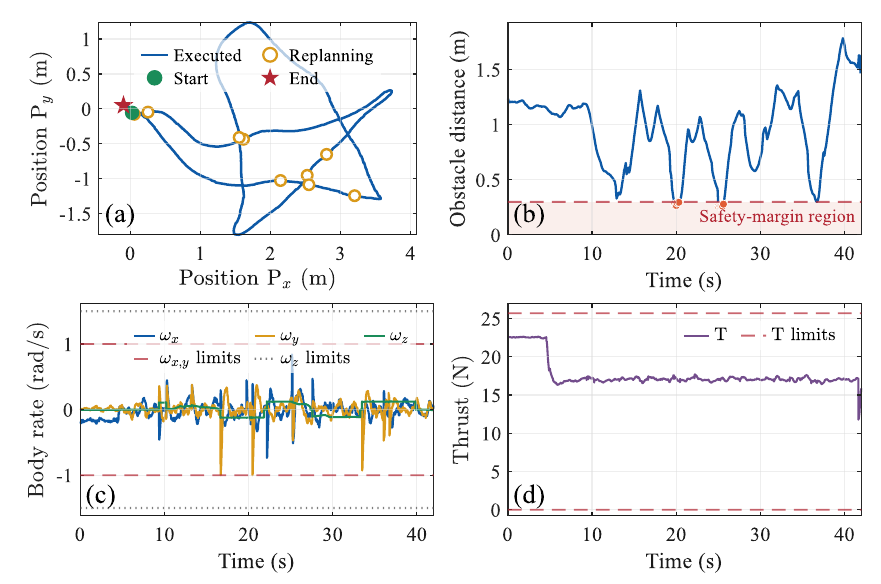}
    \caption{{DR-IPC under a dynamic obstacle}. {\footnotesize (a) trajectory and replanning; (b) nearest obstacle distance; (c) angular velocity; (d) thrust.}}
    \label{dynamic_obstacle_constraints}
\end{figure}

\begin{figure}[t]
    \centering
    \includegraphics[scale=0.60]{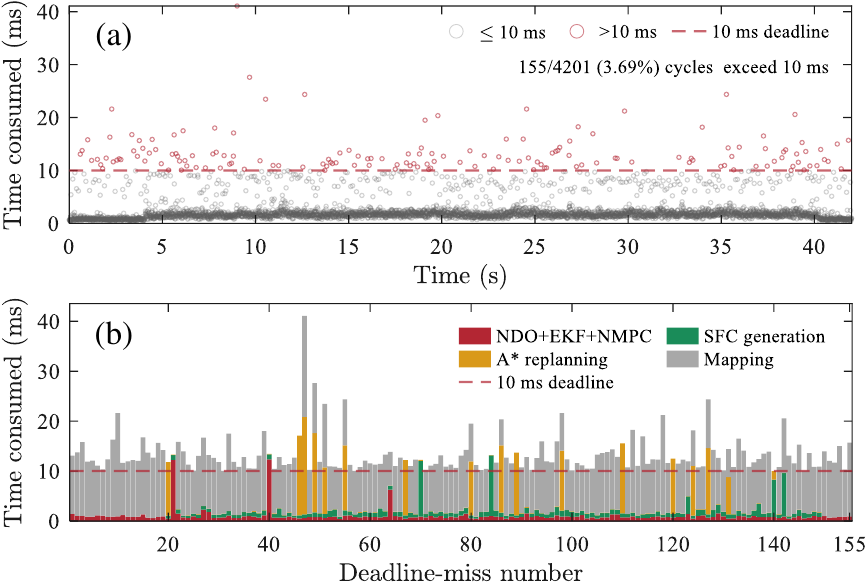}
\caption{{Computation time.} {\footnotesize (a)  per-cycle computation time; (b) breakdown of deadline-miss events.}}
    \label{figure_indoor3_plot}
\end{figure}

This experiment evaluates the complete DR-IPC pipeline in a disturbed environment with a dynamic obstacle (Unitree Go2). The continuously moving Go2 robot imposes a heavy computational load, repeatedly triggering online mapping, A*-based replanning, and SFC reconstruction, while the EKF+NDO and NMPC operate continuously. 
Representative flight snapshots, navigation and control responses, and computation times are presented in Figs.~\ref{figure_indoor3}, \ref{dynamic_obstacle_constraints}, and \ref{figure_indoor3_plot}, respectively.

As shown in Fig.~\ref{figure_indoor3}, both IPC and DR-IPC complete the task. 
However, IPC experiences multiple collisions. Fig.~\ref{dynamic_obstacle_constraints}(a) shows the executed trajectory and replanning of DR-IPC. 
Fig.~\ref{dynamic_obstacle_constraints}(b) and Fig.~\ref{figure_indoor3} show that the trajectory briefly enters the $0.30$~m inflated safety margin without collision. 
The soft SFC penalty allows this brief deviation without forcing an aggressive correction.
Meanwhile, the angular velocity and thrust commands  of DR-IPC remain within prescribed limits, as shown in Figs.~\ref{dynamic_obstacle_constraints}(c) and  (d). 

Fig.~\ref{figure_indoor3_plot} shows that 4046 of 4201 control cycles ($96.31\%$) meet the 10-ms deadline,  with 95th-percentile, 99th-percentile, and maximum computation times of 9.13, 13.71, and 41.08~ms, respectively.
Most overruns occur during map updates and A* replanning.
Among the 155 deadline misses, 153 are isolated single-cycle overruns, while the remaining two occur in consecutive cycles.
These results support soft real-time onboard execution  at 100 Hz with occasional overruns.

\begin{figure*}[t]
    \centering
    \includegraphics[scale=1.6]{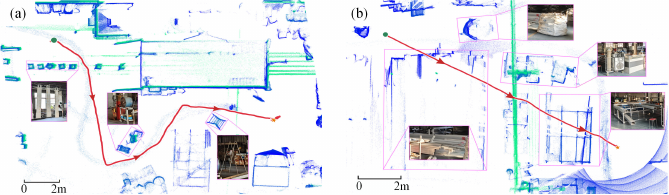}
       \caption{{LiDAR point-cloud maps and payload-delivery routes} in the factory.
    {\footnotesize (a) factory route 1; (b) factory route 2. Red curves denote the executed trajectories, green circles and orange stars mark the loading and delivery points, respectively, and inset images show representative obstacles.}}
    \label{factory_maps}
    %\label{ball_plot}
\end{figure*}

\subsection{Outdoor Experiments}
Four field scenarios are considered: daytime and nighttime flights in the same wooded area, and suspended-payload delivery along two factory routes, as shown in Fig.~\ref{outdoor1}. The wooded site contains dense vegetation, irregular branches, and narrow clearances, while nighttime operation provides limited visual cues. In the factory, machinery and stored materials restrict ground access and wheeled transport, motivating aerial delivery of lightweight payloads.

Fig.~\ref{factory_maps} shows the LiDAR point-cloud maps and executed trajectories for the two factory routes. The quadrotor transports a suspended payload from the loading point to the delivery point while avoiding columns, equipment, and stored materials. Throughout the four routes, LiDAR-based odometry and mapping, A* planning, SFC construction, EKF+NDO estimation, and NMPC operate online. Despite ambient airflow and payload-induced disturbances, DR-IPC completes all routes without collision.

\section{CONCLUSIONS}
This paper proposes DR-IPC for LiDAR-based quadrotor navigation. 
The nonlinear interconnected EKF+NDO suppresses noise and improves the accuracy and smoothness of disturbance estimation, thereby reducing fluctuations in the control commands and attitude response.
To accommodate disturbance-induced deviations, the soft SFC penalizes corridor violations while avoiding excessive corrective responses.
Together with an obstacle- and disturbance-aware NMPC backend that jointly optimizes local motion and angular-velocity/thrust commands, these components enable DR-IPC to operate in cluttered, noisy, and disturbance-rich environments. 
Comparative simulations and experiments demonstrate improved navigation reliability and disturbance rejection over IPC, while wooded and factory flights verify onboard operation in unstructured environments.

%\section{CONCLUSIONS}
%只是陈述流程，没有输出结论
% This paper proposes DR-IPC, a disturbance-resilient integrated planning and control algorithm for LiDAR-based quadrotor navigation in cluttered, noisy, and disturbance-rich environments. 
% A nonlinear interconnected observer  provides noise-suppressed state and lumped disturbance estimates for NMPC prediction. 
% %
% The soft SFC residual provides flexibility when disturbances or map updates drive the predicted motion outside the corridor.
% %
% By an obstacle- and disturbance-aware NMPC backend, DR-IPC jointly handles local path references, soft SFC residual penalties, nonlinear quadrotor dynamics, input constraints, and reconstructed disturbances  within a receding-horizon optimization.
% %
% Comparative Gazebo and indoor experiments demonstrate improved navigation reliability and disturbance rejection over IPC, while MARSIM, wooded and factory flights validate the complete system across structured and unstructured environments. 
% %
% Future work will focus on reducing nonlinear optimization cost, improving disturbance prediction under rapidly varying aerodynamic effects, and extending DR-IPC to more aggressive flight and dense dynamic environments.

% \section*{ACKNOWLEDGMENT}

%%%%%%%%%%%%%%%%%%%%%%%%%%%%%%%%%%%%%%%%%%%%%%%%%%%%%%%%%%%%%%%%%%%%%
\bibliographystyle{./mybib/IEEEtran} % use IEEEtran.bst style
\bibliography{./mybib/IEEEabrv,./mybib/mybib}

\end{document}